\documentclass[pdflatex, sn-mathphys]{sn-jnl}% Math and Physical Sciences Reference Style

\usepackage{subcaption}

\jyear{2021}%

\theoremstyle{thmstyleone}%
\newtheorem{theorem}{Theorem}%  meant for continuous numbers
\newtheorem{proposition}[theorem]{Proposition}% 
\theoremstyle{thmstyletwo}%
\theoremstyle{thmstylethree}%
\newtheorem{definition}{Definition}%
\usepackage{comment}
\begin{document}

%\title[Dual-advantage weighted supervised lea for goal-conditioned policy learning in off-line settings]{Dual-advantage weighting for goal-conditioned policy learning in off-line settings}

\title[Goal-conditioned Offline RL through State Space Partitioning]{Goal-conditioned Offline Reinforcement Learning through State Space Partitioning}

\author[1]{\fnm{Mianchu} \sur{Wang}}\email{mianchu.wang@warwick.ac.uk}
\author[1]{\fnm{Yue} \sur{Jin}}\email{yue.jin.3@warwick.ac.uk}
\author[1,2]{\fnm{Giovanni} \sur{Montana}}\email{g.montana@warwick.ac.uk\footnote{Corresponding author.} }

\affil[1]{\orgname{University of Warwick}, \orgaddress{\city{Coventry}, \country{UK}}}
\affil[2]{\orgname{Alan Turing Institute}, \orgaddress{\city{London}, \country{UK}}}

\abstract{

Offline reinforcement learning (RL) aims to create policies for sequential decision-making using exclusively offline datasets. This presents a significant challenge, especially when attempting to accomplish multiple distinct goals or outcomes within a given scenario while receiving sparse rewards. Prior methods using advantage weighting for offline goal-conditioned learning improve policies monotonically. However, they still face challenges from distribution shift and multi-modality that arise due to conflicting ways to reach a goal. This issue is especially challenging in long-horizon tasks, where the presence of multiple, often conflicting, solutions makes it hard to identify a single optimal policy for transitioning from a state to a desired goal. To address these challenges, we introduce a complementary advantage-based weighting scheme that incorporates an additional source of inductive bias. Given a value-based partitioning of the state space, the contribution of actions expected to lead to target regions that are easier to reach, compared to the final goal, is further increased. Our proposed approach, Dual-Advantage Weighted Offline Goal-conditioned RL (DAWOG), outperforms several competing offline algorithms in widely used benchmarks. Furthermore, we provide a theoretical guarantee that the learned policy will not be inferior to the underlying behavior policy.}


\maketitle

\section{Introduction}

Goal-conditioned reinforcement learning (GCRL) aims to learn policies capable of reaching a wide range of distinct goals, effectively creating a vast repertoire of skills \cite{liu2022goal, plappert2018multi, andrychowicz2017hindsight}. When extensive historical training datasets are available, it becomes possible to infer decision policies that surpass the unknown behavior policy (i.e., the policy that generated the data) in an offline manner, without necessitating further interactions with the environment \cite{eysenbach2022contrastive, mezghani2022learning, chebotar2021actionable}. A primary challenge in GCRL lies in the reward signal's sparsity: an agent only receives a reward when it achieves the goal, providing a weak learning signal. This becomes especially challenging in long-horizon problems where reaching the goals by chance alone is difficult.

In an offline setting, the challenge of learning with sparse rewards becomes even more complex due to the inability to explore beyond the already observed states and actions. When the historical data comprises expert demonstrations, imitation learning presents a straightforward approach to offline GCRL \cite{ghosh2021learning,emmons2022rvs}: in goal-conditioned supervised learning (GCSL), offline trajectories are iteratively relabeled, and a policy learns to imitate them directly. Furthermore, GCSL's objective lower bounds a function of the original GCRL objective \cite{yang2022rethinking}. However, in practice, the available demonstrations can often contain sub-optimal examples, leading to inferior policies. A simple yet effective solution involves re-weighting the actions during policy training within a likelihood maximization framework. A parameterized advantage function is employed to estimate the expected quality of an action conditioned on a target goal, so that higher-quality actions receive higher weights (Yang et al., 2022). This method is known as goal-conditioned exponential advantage weighting (GEAW).

%\footnote{\textcolor{red}{In the original paper, we used WGCSL to refer this algorithm, however it is not accurate as WGCSL consists of three parts:  discounted relabeling weighting (DRW), goal-conditioned exponential advantage weighting (GEAW), and best-advantage weighting (BAW).}}.}

Although GEAW is effective, we contend in this paper that it grapples with the pervasive multi-modality issue, especially in tasks with extended horizons. The challenge lies in pinpointing an optimal policy to achieve any set goal, given the multiple, sometimes conflicting, paths leading to that goal. While a goal-conditioned advantage function emphasizes actions likely to achieve the goal during training, we believe that introducing an extra layer of inductive bias can offer a shorter learning horizon, a robust learning signal, and more achievable objectives. This, in turn, aids the policy in discerning and adopting the best short-term trajectories amidst conflicting ones.

\begin{figure}[t]
    \centering
    \includegraphics[width=\textwidth]{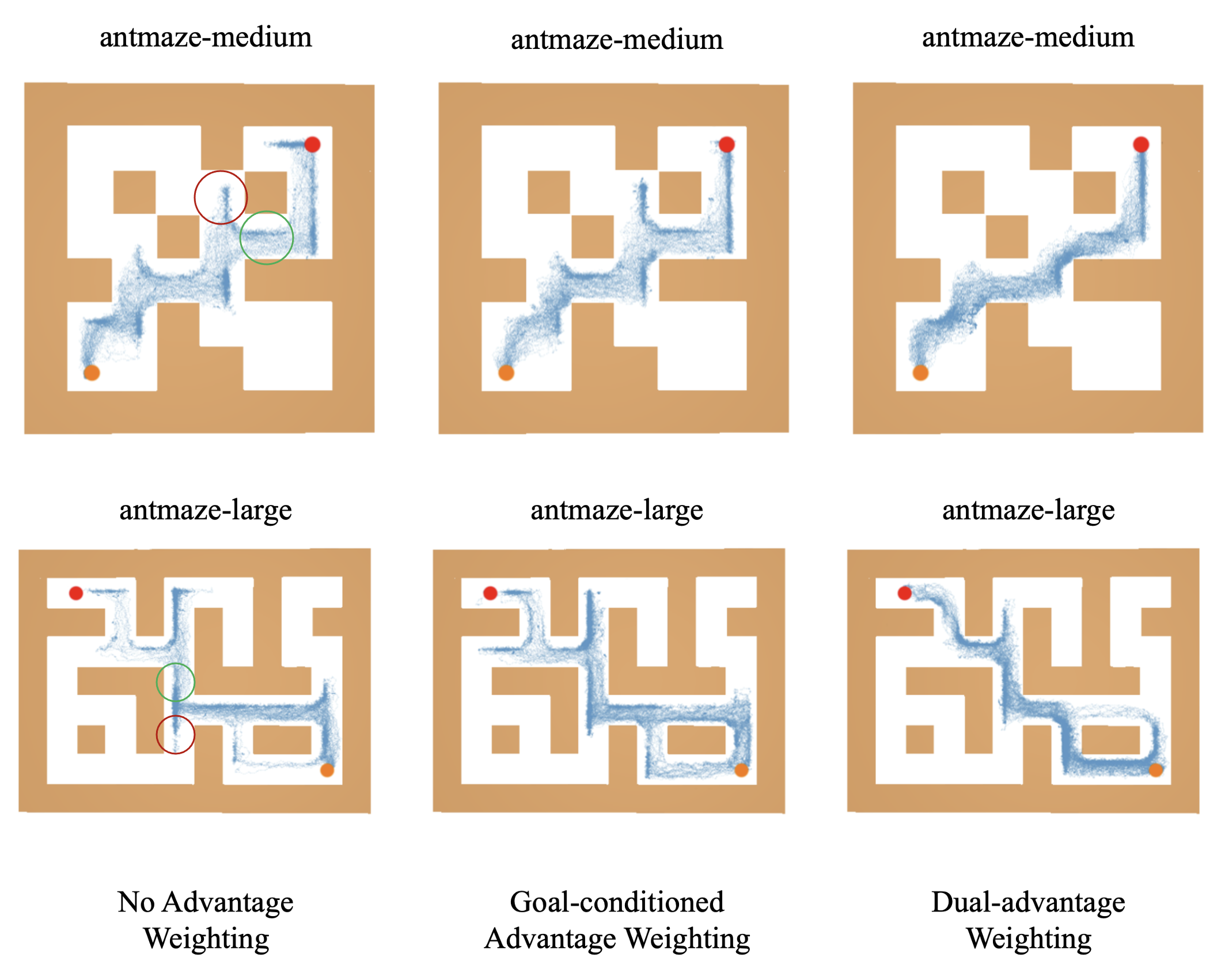}
  \caption{Visualization of trajectories (in blue) across various maze environments. These trajectories are produced by policies trained through supervised learning using different action weighting schemes: no action weighting (left), goal-conditioned advantage weighting (middle), and dual-advantage weighting (right). The task involves an agent (represented as an ant) navigating from a starting position (orange circle) to an end goal (red circle). Branching points near the circles highlight areas where the multi-modality issue is pronounced. Our proposed dual-advantage weighting scheme significantly mitigates this issue. The green circle indicates the optimal path, while the red circle marks a suboptimal route.}
    \label{fig:traj_visual_intro}
\end{figure}

\begin{figure}[t]
    \centering
    \includegraphics[width=0.47\textwidth]{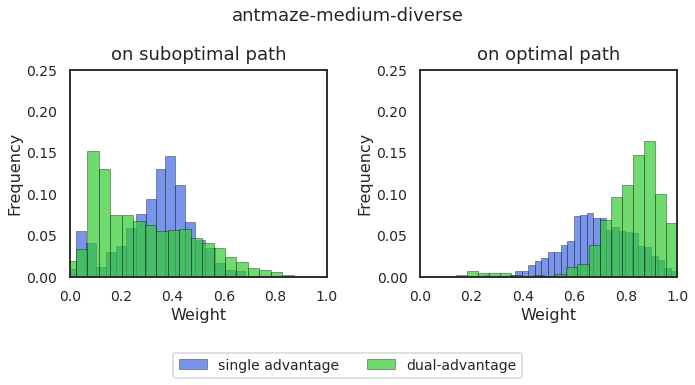}
    \includegraphics[width=0.47\textwidth]{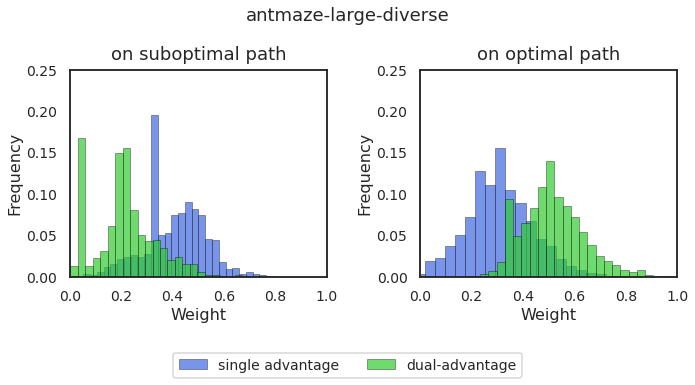}
    \caption{Comparison of normalized weights from various weighting schemes. Referring to Figure \ref{fig:traj_visual_intro}, the red circles demarcate optimal and sub-optimal areas given the target. The histograms in this figure illustrate that the dual-advantage scheme more effectively differentiates states in the optimal area from those in the sub-optimal area, allocating higher weights to the 'optimal' area states.}
    \label{fig:weight_histograms}
\end{figure}

We propose a complementary advantage weighting scheme that also utilizes the goal-conditioned value function. This provides additional guidance to address multi-modality. During training, the state space is divided into a fixed number of regions, ensuring that all states within the same region have approximately the same goal-conditioned value. These regions are then ranked from the lowest to the highest value. Given the current state, the policy is encouraged to reach the immediately higher-ranking region, relative to the state's present region, in the fewest steps possible. This \textit{target region} offers a state-dependent, short-horizon objective that is easier to achieve compared to the final goal, leading to generally shorter successful trajectories. Our proposed algorithm, Dual-Advantage Weighted Offline GCRL (DAWOG), seamlessly integrates the original goal-conditioned advantage weight with the new target-based advantage to effectively address the multi-modality issue.

A prime example is showcased in Figure \ref{fig:traj_visual_intro}, depicting the performance of three pre-trained policies in maze-based navigation tasks \cite{fu2020d4rl}. A quadruped robot has been trained to navigate these mazes. It's tasked with reaching new, unseen goals (red circles) from a starting point (orange circles). These policies were trained via supervised learning: a baseline with no action weighting (left), goal-conditioned advantage weighting (middle), and our proposed dual-advantage weighting (right). While the goal-conditioned advantage weighting often outperforms the baseline, it can occasionally guide the robot into sub-optimal areas, causing delays before redirecting towards the goal. A closer look, as shown in Figure \ref{fig:weight_histograms}, indicates that dual-advantage weighting better distinguishes goal-aligned actions from sub-optimal ones by assigning them different weights. Consequently, our dual-advantage weighting approach mitigates the multi-modality challenge, resulting in policies that offer more direct and efficient routes to the goal.

In this work, we address the challenges of multi-modality in goal-conditioned offline RL, introducing a novel approach to tackle them. The main contributions of our paper are:
\begin{enumerate}
    \item A proposed dual-advantage weighted supervised learning approach, tailored to mitigate the multi-modality challenges inherent in goal-conditioned offline RL.
    \item Theoretical assurances that our method’s performance matches or exceeds that of the underlying behavior policy.
    \item Empirical evaluations across diverse benchmarks \cite{fu2020d4rl, plappert2018multi, yang2022rethinking} showcasing DAWOG’s consistent edge over other leading algorithms.
    \item A series of studies highlighting DAWOG's unique properties and its robustness against hyperparameter variations.
\end{enumerate}

\section{Related work}

In this section, we offer a brief overview of methodologically related approaches. In goal-conditioned RL (GCRL) In \textbf{goal-conditioned RL (GCRL)}, one of the main challenges is the sparsity of the reward signal. An effective solution is hindsight experience replay (HER) \cite{andrychowicz2017hindsight}, which relabels failed rollouts that have not been able to reach the original goals and treats them as successful examples for different goals thus effectively learning from failures. HER has been extended to solve different challenging tasks in synergy with other learning techniques, such as curriculum learning \cite{fang2019curriculum}, model-based goal generation \cite{yang2021mher, jurgenson2020sub, nasiriany2019planning, nair2018visual}, and generative adversarial learning \cite{durugkar2021adversarial,charlesworth2020plangan}. In the offline setting, GCRL aims to learn goal-conditioned policies using only a fixed dataset. The simplest solution has been to adapt standard offline reinforcement learning algorithms \cite{kumar2020conservative,fujimoto2021minimalist} by simply concatenating the state and the goal as a new state. Chebotar \textit{et al.} \cite{chebotar2021actionable} propose goal-conditioned conservative Q-learning and goal-chaining to prevent value over-estimation and increase the diversity of the goal. Some of the previous works design offline GCRL algorithms from the perspective of state-occupancy matching \cite{eysenbach2022contrastive}. Mezghani \textit{et al.} \cite{mezghani2022learning} propose a self-supervised reward shaping method to facilitate offline GCRL.

Our work is most related to \textbf{goal-conditioned imitation learning (GCIL)}.   Emmons \textit{et al.} \cite{emmons2022rvs} study the importance of concatenating goals with states showing its effectiveness in various environments. 
Ding \textit{et al.}  \cite{ding2019goal} extend generative adversarial imitation learning \cite{ho2016generative} to goal-conditioned settings. 
Ghosh \textit{et al.} \cite{ghosh2021learning} extend behaviors cloning \cite{bain1995framework} to goal-conditioned settings and propose goal-conditioned supervised learning (GCSL) to imitate relabeled offline trajectories. 
Yang \textit{et al.} \cite{yang2022rethinking} connect GCSL to offline GCRL, and show that the objective function in GCSL is a lower bound of a function of the original GCRL objective. They propose the GEAW algorithm, which re-weights the offline data based on advantage function similarly to \cite{peng19advantage,wang2018exponentially}. Additionally, Yang \textit{et al.} \cite{yang2022rethinking} identify the multi-modality challenges in GEAW and introduce the best-advantage weight (BAW) to exclude state-actions with low advantage during the learning process. In parallel, our DAWOG was developed to address this very challenge, offering a novel advantage-based action re-weighting approach.

Some connections can also be found with \textbf{goal-based hierarchical reinforcement learning} methods\cite{li2022hierarchical,chane2021goal,kim2021landmarkguided,zhang2021worldmodel,nasiriany2019planning}. These works feature a high-level model capable of predicting a sequence of intermediate sub-goals and learn low-level policies to achieve them. Instead of learning to reach a specific sub-goals, our policy learns to reach an entire sub-region of the state space containing states that are equally valuable and provide an incremental improvement towards the final goal. 
%In a way, our approach  inherits the advantage of hierarchical reinforcement learning in long horizon tasks while avoiding the out-of-distribution goals generated from the high-level model.

Lastly, there have been other applications of \textbf{state space partitioning} in reinforcement learning, such as facilitating exploration and accelerating policy learning in online settings \cite{ma2020clustered,wei2018learning,karimpanal2017identification,mannor2004dynamic}. Ghosh \textit{et al.} \cite{ghosh2018divide} demonstrate that learning a policy confined to a state partition instead of the whole space can lead to low-variance gradient estimates for learning value functions.  In their work, states are partitioned using K-means to learn an ensemble of locally optimal policies, which are then progressively merged into a single, better-performing policy. Instead of partitioning states based on their geometric proximity, we partition states according to the proximity of their corresponding goal-conditioned values. We then use this information to define an auxiliary reward function and, consequently, a region-based advantage function.

\section{Preliminaries} \label{sec:preliminaries}
\textbf{Goal-conditioned MDPs.} Goal-conditioned tasks are usually modeled as Goal-Conditioned Markov Decision Processes (GCMDP), denoted by a tuple $<\mathcal{S}, \mathcal{A}, \mathcal{G}, P, R>$ where $\mathcal{S}$, $\mathcal{A}$, and $\mathcal{G}$ are the state, action and goal space, respectively. For each state $s \in \mathcal{S}$, there is a corresponding achieved goal, $\phi(s) \in \mathcal{G}$, where $\phi:\mathcal{S} \rightarrow \mathcal{G}$ \cite{liu2022goal}. At a given state $s_t$, an action $a_t$  taken towards a desired goal $g$ results in a visited next state $s_{t+1}$ according to the environment's transition dynamics, $P(s_{t+1} \mid s_t, a_t)$. The environment then provides a reward, $r_t = R(s_{t+1}, g)$, which is non-zero  only when the goal has been reached, i.e.,
\begin{equation} \label{reward_function}
    R(s, g) =
    \begin{cases}
      1, & \text{if $\mid \mid \phi(s) - g \mid \mid ^2_2 \leq \text{threshold}$,}\\
      0, & \text{otherwise.}\\
    \end{cases}       
\end{equation}

\textbf{Offline Goal-conditioned RL.} In offline GCRL, the agent aims to learn a goal-conditioned policy, $\pi: \mathcal{S} \times \mathcal{G} \rightarrow \mathcal{A}$, using an offline dataset containing previously logged trajectories that might be generated by any number of unknown behaviors policies. 
% The historical data might have been collected by any number of unknown behavior policies. 
The objective is to maximize the expected and discounted cumulative returns, 
\begin{equation} \label{rl_objective}
    J_{GCRL}(\pi)=\mathbb{E}_{\substack{g \sim P_g, s_0 \sim P_0, \\a_t \sim \pi(\cdot \mid s_t, g), \\s_{t+1} \sim P(\cdot \mid s_t, a_t)}} \left [ \sum^T_{t=0} \gamma^t r_t \right ],
\end{equation}
where $\gamma \in (0, 1]$ is a discount factor, $P_g$ is the distribution of the goals, $P_0$ is the distribution of the initial state, and $T$ 
% is the terminal time step; 
corresponds to the time step at which an episode ends, i.e., either the goal has been achieved or timeout has been reached. 

\textbf{Goal-conditioned Value Functions.} A goal-conditioned state-action value function \cite{schaul2015universal} quantifies the value of an action $a$ taken from a state $s$ conditioned on a goal $g$ using the sparse rewards of Eq. \ref{reward_function},
\begin{equation} \label{eq:goal_q}
    Q^\pi(s, a, g)=\mathbb{E}_{\pi} \left [ \sum^T_{t=0} \gamma^t r_t \mid s_0=s, a_0=a \right ]
\end{equation}
where $\mathbb{E}_{\pi}[\cdot]$ denotes the expectation taken with respect to $a_t \sim \pi(\cdot \mid s_t, g)$ and $s_{t+1} \sim P(\cdot \mid s_t, a_t)$. Analogously, the goal-conditioned state value function quantifies the value of a state $s$ when trying to reach $g$,
\begin{equation} \label{pre:goal_v}
    V^\pi(s, g)=\mathbb{E}_{\pi} \left [ \sum^T_{t=0} \gamma^t r_t \mid s_0=s\right ].
\end{equation}
The goal-conditioned advantage function, 
\begin{equation} \label{eq:goal_adv}
    A^\pi(s, a, g)=Q^\pi(s, a, g)-V^\pi(s, g),
\end{equation}
then quantifies how advantageous it is to take a specific action $a$ in state $s$ towards $g$ over taking the actions sampled from $\pi(\cdot \mid s, g)$ \cite{yang2022rethinking}. 

\textbf{Goal-conditioned Supervised Learning (GCSL).} GCSL \cite{ghosh2021learning} relabels the desired goal in each data tuple $(s_t, a_t, g)$ with the goal achieved henceforth in the trajectory to increase the diversity and quality of the data \cite{andrychowicz2017hindsight, kaelbling1993learning}. The relabeled dataset is denoted as 
$\mathcal{D}_R=\{(s_t, a_t, g=\phi(s_i)) \mid T \geq i > t \geq 0\}$. GCSL 
% trains decision policies 
learns a policy that mimics the relabeled transitions by maximizing
\begin{equation} \label{eq:gcsl_obj}
    J_{GCSL}(\pi) = \mathbb{E}_{(s_t, a_t, g) \sim \mathcal{D}_R} \left [ \pi(a_t \mid s_t, g) \right ].
\end{equation}
Yang \textit{et al.} \cite{yang2022rethinking} have connected GCSL to GCRL and demonstrated that $J_{GCSL}$ lower bounds $\frac{1}{T}\log J_{GCRL}$.

\textbf{Goal-conditioned Exponential Advantage Weighting (GEAW).} GEAW, as discussed in \cite{yang2022rethinking, wang2018exponentially}, extends GCSL by incorporating a goal-conditioned exponential advantage as the weight for Eq. \ref{eq:gcsl_obj}. Its design ensures that samples with higher advantages receive larger weights and vice versa. Specifically, GEAW trains a policy that emulates relabeled transitions, but with varied weights:
\begin{equation}
J_{GEAW}(\pi) = \mathbb{E}_{(s_t, a_t, g) \sim \mathcal{D}_R} \left [ \exp_{clip} (A(s_t, a_t, g)) \pi(a_t \mid s_t, g) \right ].
\end{equation}
Here, $\exp_{clip}(\cdot)$ clips values within the range $(0, M]$ to ensure numerical stability. This weighting approach has been demonstrated as a closed-form solution to an offline RL problem, guaranteeing that the resultant policy aligns closely with the behavior policy \cite{wang2018exponentially}.

\section{Methods} \label{sec:methods}

In this section, we  formally present the proposed methodology and analytical results. First, we introduce a notion of target region advantage function in Section \ref{function}, which we use to develop the learning algorithm in Section \ref{algo}.  In Section \ref{theory} we provide a theoretical analysis offering guarantees that  DAWOG  learns a policy that is never worse than the underlying behaviors policy.

%\subsection{Overview}\label{overview}

%We propose dual-advantage weighted offline GCRL (DAWOG) algorithm to solve GCRL problems. DAWOG utilises an exponential weighting factor that combines the goal-conditioned advantage function and a novel region-based advantage function to incentivize actions to reach the goal  within as few time steps as possible. 

% In the reminder of this section, we properly define the target region, define the target region-conditioned value functions and the advantage function, and present the full algorithm. Finally, we demonstrate that the policy learned by the algorithm is not worse than the behavioral policy that generates the relabeled data. 

\begin{figure}[t]
    \centering
    \includegraphics[width=\textwidth]{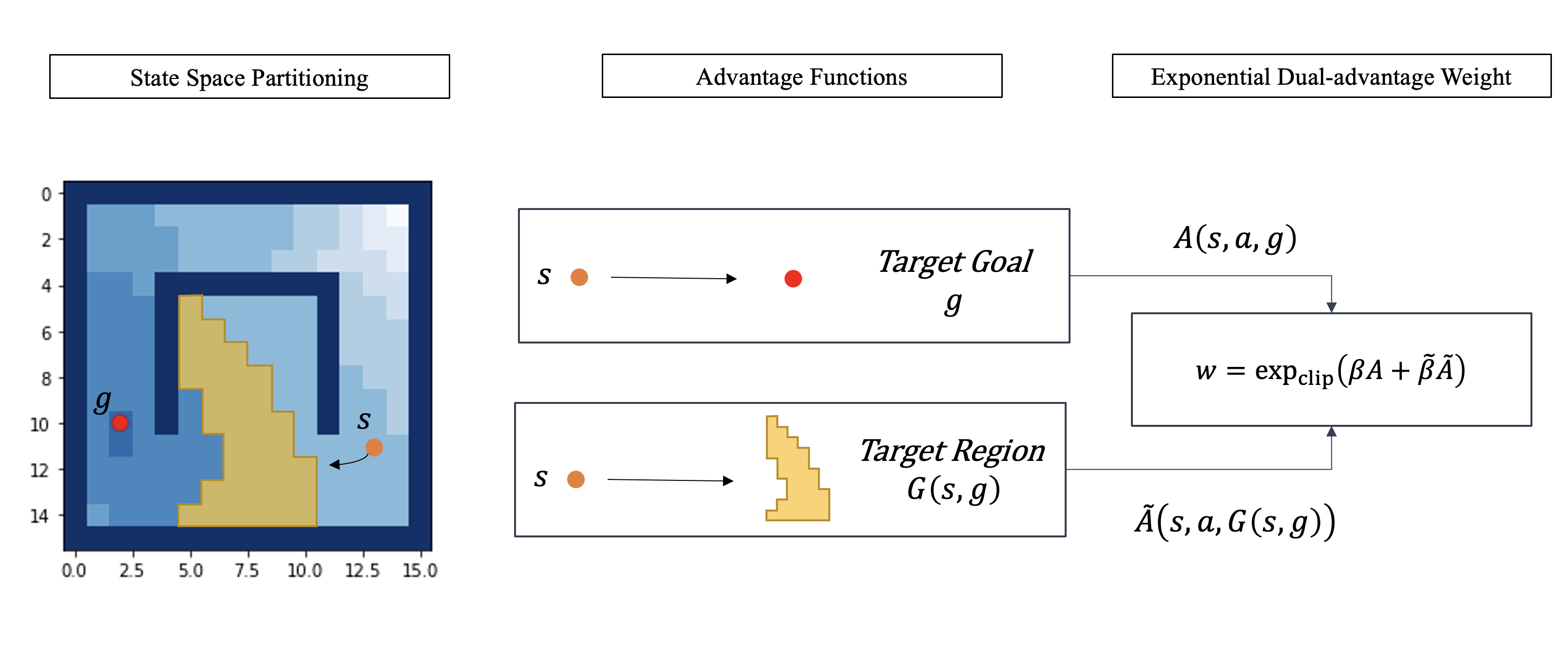}
    \caption{Illustration of the two advantage functions used by DAWOG for a simple navigation task. First, a goal-conditioned advantage is learned using only relabeled offline data. Then, a target-region advantage is obtained by partitioning the states according to their goal-conditioned value function, identifying a target region, and rewarding actions leading to this region in the smallest possible number of steps. DAWOG updates the policy to imitate the offline data through an exponential weighting factor that depends on both advantages.}
   \label{fig:algo_diagram}
\end{figure}

\subsection{Target region advantage function} \label{function}

For any state $s \in \mathcal{S}$ and goal $g \in \mathcal{G}$, the domain of the goal-conditioned value function in Eq. \ref{pre:goal_v} is the unit interval due to the binary nature of the reward function in Eq. \ref{reward_function}. Given a positive integer $K$, we partition $[0, 1]$ into $K$ equally sized intervals, $\{ \beta_i \}_{i=1,\ldots,K}$. For any goal $g$, this partition induces a corresponding partition of the state space. 

\begin{definition} \label{def:bin}
(Goal-conditioned State Space Partition)
For a fixed desired goal $g \in \mathcal{G}$, the state space is partitioned into $K$ equally sized regions according to $V^\pi(\cdot, g)$. The $k^{th}$ region, notated as $B_{k}(g)$, contains  all states whose goal-conditioned values are within $\beta_k$, i.e.,
\begin{equation}
B_{k} (g) = \{s \in \mathcal{S} \mid  V^\pi(s, g) \in \beta_k \}. 
\end{equation}
\end{definition}

Our ultimate objective is to up-weight actions taken in a state $s_t \in B_{k}(g)$ that are likely to lead to a region only marginally better (but never worse) than $B_{k}(g)$ as rapidly as possible. 

\begin{definition} \label{def:target_region}
(Target Region) For $s \in B_{k}(g)$, the mapping $b(s, g): \mathcal{S} \times \mathcal{G} \rightarrow \{1, \ldots, K\}$ returns the correct index $k$. The goal-conditioned target region is defined as 
\begin{equation}
    G(s, g) = B_{\min\{b(s, g)+1, K\}}(g),
\end{equation}
which is the set of states whose goal-conditioned value is not less than the states in the current region. For $s \in B_k(g)$, $G(s, g)$ is the current region $B_k(g)$ if and only if $k=K$. 
\end{definition}

We now introduce two target region value functions. 

\begin{definition} (Target Region Value Functions) For a state $s$, action $a \in \mathcal{A}$, and the target region $G(s, g)$, we define a target region V-function and a target region Q-function based on an auxiliary reward function that returns a non-zero reward only when the next state belongs to the target region, i.e., 
\begin{equation}
    \tilde{r}_t = \tilde{R}(s_t, s_{t+1}, G(s_t, g)) = 
    \begin{cases}
      1, & \text{if } s_{t+1} \in G(s_t,g)\\
      0, & \text{otherwise.} \\
    \end{cases}
\end{equation}
The target region Q-value function is 
\begin{equation}
    \Tilde{Q}^\pi(s, a, G(s, g)) = \mathbb{E}_{\pi} \left [ \sum^{\tilde{T}}_{t=0} \gamma^t \tilde{r}_t \mid s_0=s, a_0=a \right ],
\end{equation}
where $\tilde{T}$ corresponds to the time step at which the target region is achieved or timeout is reached, $\mathbb{E}_{\pi}[\cdot]$ denotes the expectation taken with respect to the policy $a_t \sim \pi(\cdot \mid s_t, g)$ and the transition dynamics $s_{t+1} \sim P(\cdot \mid s_t, a_t)$. The target region Q-function estimates the expected cumulative return when starting in $s_t$, taking an action $a_t$, and then following the policy $\pi$, based on the auxiliary reward. The discount factor $\gamma$ reduces the contribution of delayed target achievements. Analogously, the target region value function is defined as
\begin{equation}
    \Tilde{V}^\pi(s, G(s, g)) = \mathbb{E}_{\pi} \left [ \sum^{\tilde{T}}_{t=0} \gamma^t \tilde{r}_t \mid s_0=s \right ]
\end{equation}
and quantifies the quality of a state $s$ according to the same criterion. 
\end{definition}

Using the above value functions, we are in a position to introduce the corresponding target region advantage function. 
\begin{definition} (Target Region Advantage Function)
    The target region-based advantage function is defined as 
    \begin{equation} \label{eq:region_adv} 
    \tilde{A}^\pi(s, a, G(s,g)) =  \tilde{Q}^\pi(s, a, G(s, g))  - \tilde{V}^\pi(s, G(s, g)).
    \end{equation}
    It estimates the advantage of action $a$ towards the target region in terms of the cumulative return by taking $a$ in state $s$ and following the policy $\pi$ thereafter, compared to taking actions sampled from the policy.
\end{definition}

\subsection{The DAWOG algorithm}\label{algo}

The proposed DAWOG belongs to the family of WGCSL algorithms, i.e. it is designed to optimize the following objective function 
\begin{equation} \label{eq:policy}
    J_{DAWOG}(\pi) = \mathbb{E}_{(s_t, a_t, g) \sim \mathcal{D}_R} \left [ w_t \log \pi(a_t \mid s_t, g) \right ]
\end{equation}
where the role of $w_t$ is to re-weight each action's contribution to the loss. In DAWOG, $w_t$ is an exponential weight of form 
\begin{equation}\label{eq:weight}
    w_t = \exp_{clip} (\beta A^{\pi_b}(s_t, a_t, g) + \Tilde{\beta} \Tilde{A}^{\pi_b}(s_t, a_t, G(s_t, g)),
\end{equation}
where $\pi_b$ is the underlying behavior policy that generate the relabeled dataset $\mathcal{D}_R$. The contribution of the two advantage functions, $\Tilde{A}^{\pi_b}(s_t, a_t, g)$ and $\Tilde{A}^{\pi_b}(s_t, a_t, G(s_t,g))$, is controlled by positive scalars, $\beta$ and $\Tilde{\beta}$, respectively. However, empirically, we have found that using a single shared parameter generally performs well across the tasks we have considered (see Section \ref{further_studies}). The clipped exponential, $\exp_{clip}(\cdot)$, is used for numerical stability and keeps the values within the $(0, M]$ range, for a given $M>0$  threshold.

The algorithm combines the originally proposed goal-conditioned advantage \cite{yang2022rethinking} with the novel target region advantage. The former ensures that actions likely to lead to the goal are up-weighted. However, when the goal is still far, there may still be several possible ways to reach it, resulting in a wide variety of favorable actions. The target region advantage function provides additional guidance by further increasing the contribution of actions expected to lead to a higher-valued sub-region of the state space as rapidly as possible. Both $A^{\pi_b}(s_t, a_t, g)$ and $\Tilde{A}^{\pi_b}(s_t, a_t, G(s_t,g))$ are beneficial in a complementary fashion: whereas the former is more concerned with long-term gains, which are more difficult and uncertain, the latter is more concerned with short-term gains, which are easier to achieve. As such, these two factors are complementary and their combined effect plays an important role in the algorithm's final performance (see Section \ref{further_studies}).  An illustration of the dual-advantage weighting scheme is shown in Fig.\ref{fig:algo_diagram}. 

In the remainder, we explain the entire training procedure. The advantage $A^{\pi_b}(s_t, a_t, g)$ is estimated via 
\begin{equation}\label{goal-advantage}
A^{\pi_b}(s_t, a_t, g)= r_t + \gamma V^{\pi_b}(s_{t+1}, g) - V^{\pi_b}(s_t, g).
\end{equation} 

In practice, the goal-conditioned V-function is approximated by a deep neural network with parameter $\psi_1$, which is learned by minimizing the temporal difference (TD) error \cite{sutton2018reinforcement}:
\begin{equation} \label{eq:goal_value}
    \mathcal{L}(\psi_1) = \mathbb{E}_{(s_t, s_{t+1}, g) \sim \mathcal{D}_R} \left [ (V_{\psi_1}(s_t, g) - y_t)^2 \right ],
\end{equation}
where $y_t$ is the target value given by
\begin{equation}
    y_t=r_t + \gamma (1 - d(s_{t+1}, g)) V_{\psi^-_1}(s_{t+1}, g).
\end{equation}
Here $d(s_{t+1}, g)$) indicates whether the state $s_{t+1}$ has reached the goal $g$. The parameter vector $\psi^-_1$ is a slowly moving average of $\psi_1$ to stabilize training \cite{mnih2015human}. Analogously, the target region advantage function is estimated by 
\begin{equation}\label{region-advantage}
    \tilde{A}^{\pi_b}(s_t, a_t, G(s_t, g))= \tilde{r}_t + \gamma \tilde{V}^{\pi_b}(s_{t+1}, G(s_t, g)) - \tilde{V}^{\pi_b}(s_t, G(s_t, g)),
\end{equation}
where the target region V-function is approximated with a deep neural network parameterized with $\psi_2$. The relevant loss function is
\begin{equation} \label{eq:region_value}
    \mathcal{L}(\psi_2) = \mathbb{E}_{(s_t, s_{t+1}, g) \sim \mathcal{D}_R} \left [ (\tilde{V}_{\psi_2}(s_t, G(s_t, g)) - \tilde{y}_t)^2 \right ],
\end{equation}
where the target value is 
\begin{equation}
    \tilde{y}_t=\tilde{r}_t + \gamma (1 - \tilde{d}(s_{t+1}, G(s_t, g))) \tilde{V}_{\psi^-_2}(s_{t+1}, G(s_t, g)).
\end{equation}
and $\tilde{d}(s_{t+1}, G(s_t, g))$) indicates whether the state $s_{t+1}$ has reached the target region $G(s_t, g)$. $\psi^-_2$ is a slowly moving average of $\psi_2$.  The full procedure is presented in Algorithm \ref{pseudocode} where the two value functions are jointly optimized and contribute to optimizing Eq. \ref{eq:policy}. 

\begin{algorithm}[t]
\caption{Dual-Advantage Weighted Offline GCRL (DAWOG)} \label{pseudocode}
\begin{algorithmic}[1]
% \textbf{Input:} relabeled dataset $\mathcal{D}_R$
\renewcommand{\algorithmicrequire}{\textbf{Initialize:}}
\Require parameters $\theta$, $\psi_1$, $\psi^-_1$, $\psi_2$, and $\psi^-_2$ for policy network, goal-conditioned value network and its target network, partition-conditioned value network and its target network, respectively; relabeled dataset $\mathcal{D}_R$.
\While{\textit{not converged}}
    \For {$i=1, \ldots,I$}
        \State Sample a minibatch $\{(s, a, s', g)\}^B_{b=1}$ from $D_R$.
        \State Update $\psi_1$ to minimize Eq. \ref{eq:goal_value}.
        \State Update $\psi_2$ to minimize Eq. \ref{eq:region_value}.
        \State Comp. $A(s, a, g) = r_t + \gamma V_{\psi_1}(s', g) - V_{\psi_2}(s, g)$.
        \State Comp. $\tilde{A}(s, a, G(s, g)) = \tilde{r}_t + \gamma \tilde{V}_{\psi_2}(s', G(s, g)) - \tilde{V}_{\psi_2}(s, G(s, g))$.
        \State Update the policy by minimizing Eq. \ref{eq:policy}.
    \EndFor
    \State $\psi^-_1 \leftarrow \rho \psi_1 + (1-\rho) \psi^-_1$.
    \State $\psi^-_2 \leftarrow \rho \psi_2 + (1-\rho) \psi^-_2$.
\EndWhile
\State \Return $\pi_\theta$.
\end{algorithmic}
\end{algorithm}

\subsection{Policy improvement guarantees}\label{theory}
In this section, we demonstrate that our learned policy is never worse than the underlying behavior policy $\pi_b$ that generates the relabeled data. First,  we express the policy learned by our algorithm in an equivalent form, as follows. 

\begin{proposition}
    DAWOG learns a policy $\pi_\theta$ to minimize the KL-divergence from 
    \begin{equation} \label{eq:dual_imitate_policy}
        \tilde{\pi}_{dual}(a \mid s, g) = \exp (w + N(s, g)) \pi_b(a \mid s, g), 
    \end{equation}
    where $w = \beta A^{\pi_b}(s, a, g) + \tilde{\beta} \tilde{A}^{\pi_b}(s, a, G(s,g))$, $G(s,g)$ is the target region, and $N(s, g)$ is a normalizing factor to ensuring that $\sum_{a \in \mathcal{A}} \tilde{\pi}_{dual}(a \mid s, g)=1$. 
\end{proposition}
\begin{proof}
    According to Eq. \ref{eq:policy}, DAWOG maximizes the following objective with the policy parameterized by $\theta$:
    \begin{align}
    \begin{split}
        \arg \max_\theta J(\theta) = & \arg \max_\theta \mathbb{E}_{(s, a, g) \sim \mathcal{D}_R} \left [ \exp (w + N(s, g)) \log \pi_\theta(a \mid s, g)\right ] \\
        = & \arg \max_\theta \mathbb{E}_{(s, g) \sim \mathcal{D}_R} \left [ \sum_{a} \exp(w + N(s, g)) \pi_b(a \mid s, g) \log \pi_\theta(a \mid s, g) \right ] \\
        = & \arg \max_\theta \mathbb{E}_{(s, g) \sim \mathcal{D}_R} \left [  \sum_{a} \tilde{\pi}_{dual}(a \mid s, g) \log \pi_\theta(a \mid s, g) \right ] \\
        = & \arg \max_\theta \mathbb{E}_{(s, g) \sim \mathcal{D}_R} \left [  \sum_{a} \tilde{\pi}_{dual}(a \mid s, g) \log \frac{\pi_\theta(a \mid s, g)}{\tilde{\pi}_{dual}(a \mid s, g)} \right ]\\
        = & \arg \min_\theta \mathbb{E}_{(s, g) \sim \mathcal{D}_R} \left [  D_{KL} (\tilde{\pi}_{dual}(\cdot \mid s, g) \mid \mid \pi_\theta(\cdot \mid s, g)) \right ]
    \end{split}
    \end{align}
    $J(\theta)$ reaches its maximum when
    \begin{equation}
        D_{KL} (\tilde{\pi}_{dual}(\cdot \mid s, g) \mid \mid \pi_{\theta}(\cdot \mid s, g)) = 0, \forall s \in \mathcal{S}, g \in \mathcal{G}.
    \end{equation}
\end{proof}

Then, we propose Proposition \ref{prop:mono_inc} to show the condition for policy improvement.
\begin{proposition} \label{prop:mono_inc}
\textup{\cite{wang2018exponentially,yang2022rethinking}} Suppose two policies $\pi_1$ and $\pi_2$ satisfy
\begin{equation} \label{eq:mono_inc}
    h_1(\pi_2(a \mid s, g)) = h_1(\pi_1(a \mid s, g)) + h_2(s, g, A^{\pi_1}(s, a, g))
\end{equation}
where $h_1(\cdot)$ is a monotonically increasing function, and $h_2(s, g, \cdot)$ is monotonically increasing for any fixed $s$ and $g$. Then we have 
\begin{equation*}
    V^{\pi_2}(s, g) \geq V^{\pi_1}(s, g), \forall s \in \mathcal{S} \text{ and } g \in \mathcal{G}.
\end{equation*}
That is, $\pi_2$ is uniformly as good as or better than $\pi_1$.
\end{proposition}

We want to leverage this result to demonstrate that  $V^{\tilde{\pi}_{dual}}(s, g) \geq V^{\pi_b}(s, g)$ for any state $s$ and goal $g$. Firstly, we need to obtain a monotonically increasing function  $h_1(\cdot)$. This is achieved by taking the logarithm of the both sides of Eq. \ref{eq:dual_imitate_policy}, i.e.,
\begin{align}
\begin{split}
    \log \tilde{\pi}_{dual}(a \mid s, g) =& \log \pi_b(a \mid s, g) + \beta A^{\pi_b}(s, a, g) \\& + \tilde{\beta} \tilde{A}^{\pi_b}(s, a, G(s,g)) + N(s, g).
\end{split}
\end{align}
so that $h_1(\cdot)= \log(\cdot)$. The following proposition establishes that we also have a function $h_2(s, g, A^{\pi_b}(s, a, g)) =  \beta A^{\pi_b}(s, a, g) + \tilde{\beta} \tilde{A}^{\pi_b}(s, a, G(s,g)) + N(s, g)$, which is monotonically increasing for any fixed $s$ and $g$. Since $\beta, \tilde{\beta} \geq 0$ and $N(s, g)$ is independent of the action, it is equivalent to prove that for any fixed $s$ and $g$, there exists a monotonically increasing function $l$ satisfying 
\begin{equation}
l(s, g, \tilde{A}^{\pi_b}(s, a, G(s,g))) = A^{{\pi_b}}(s, a, g).
\end{equation}
% the fact can be written in the form of Proposition \ref{prop: mono_proof}.

\begin{proposition} \label{prop: mono_proof}
    Given fixed $s$, $g$ and the target region $G(s,g)$, the goal-conditioned advantage function $A^\pi$ and the target region-conditioned advantage function $\tilde{A}^\pi$ satisfy 
    $l(s, g, A^\pi(s, a, g)) = \tilde{A}^\pi(s, a, G(s,g))$, where $l(s, g,\cdot)$ is monotonically increasing for any fixed $s$ and $g$.
\end{proposition}

\begin{proof}
    By the definition of monotonically increasing function, if for all $a', a'' \in \mathcal{A}$ such that $A^\pi(s, a', g) \geq A^\pi(s, a'', g)$ and we can reach $\tilde{A}^\pi(s, a', G(s,g)) \geq \tilde{A}^\pi(s, a'', G(s,g))$, then the proposition can be proved.
    
    We start by having any two actions $a', a'' \in \mathcal{A}$ such that 
    \begin{equation}
        A^\pi(s, a', g) \geq A^\pi(s, a'', g).
    \end{equation}
    By adding $V^\pi(s, g)$ on both sides, the inequality becomes
    \begin{equation} \label{Q_ineq}
        Q^\pi(s, a', g) \geq Q^\pi(s, a'', g). 
    \end{equation}
    By Definition \ref{eq:goal_q}, the goal-conditioned Q-function can be written as
    \begin{equation} \label{Q_traj}
        Q^\pi(s, a, g)=\mathbb{E}_{\pi} \left [ R_{t,\tau_i} \mid s_t=s, a_t=a \right ],
    \end{equation}
    where $\tau_i$ represents a trajectory: $s_t, a_t, r_t^i, s_{t+1}^i, a_{t+1}^i, r_{t+1}^i, \ldots, s_{T}^i$
    \begin{equation} \label{}
        R_{t,\tau_i}=r_t^i + \gamma r_{t+1}^i + \ldots + \gamma^{t_{tar}^i}V(s_{tar}^i, g),
    \end{equation}
    $s_{tar}^i$ corresponds to the state where $\tau_i$ gets into the target region, $t_{tar}^i$ is the corresponding time step.
    Because the reward is zero until the desired goal is reached, Eq. \ref{Q_traj} can  be written as 
    \begin{equation} \label{Q_traj_}
        Q^\pi(s, a, g)=\mathbb{E}_{\pi} \left [\gamma^{t_{tar}^i}V(s_{tar}^i, g) \mid s_t=s, a_t=a \right ].
    \end{equation}
    Similarly, 
    \begin{equation} \label{our_Q_traj}
    \begin{aligned}
        \tilde{Q}^\pi(s, a, G(s,g))&=\mathbb{E}_{\pi} \left [\gamma^{t_{tar}^i}\tilde{V}(s_{tar}^i, G(s,g)) \mid s_t=s, a_t=a \right ]\\
        &=\mathbb{E}_{\pi} \left [\gamma^{t_{tar}^i} \mid s_t=s, a_t=a \right ].
    \end{aligned}
    \end{equation}
    According to Eq.\ref{Q_ineq} and Eq.\ref{Q_traj_}, we have
    \begin{equation} \label{Q_ineq_}
        \mathbb{E}_{\pi} \left [\gamma^{t_{tar}^{i'}}V(s_{tar}^{i'}, g) \mid s_t=s, a_t=a' \right ] \geq \mathbb{E}_{\pi} \left [\gamma^{t_{tar}^{i''}}V(s_{tar}^{i''}, g) \mid s_t=s, a_t=a'' \right ].
    \end{equation}
    Given the valued-based partitioning of the state space, we assume that the goal-conditioned values of states in the target region are sufficiently close such that $\forall i', i'', V(s_{tar}^{i'}, g) \approx v,  V(s_{tar}^{i''}, g) \approx v $. Then, Eq.\ref{Q_ineq_} can be approximated as 
    \begin{equation} \label{Q_ineq_final}
        v \cdot \mathbb{E}_{\pi} \left [\gamma^{t_{tar}^{i'}} \mid s_t=s, a_t=a' \right ] \geq  v \cdot \mathbb{E}_{\pi} \left [\gamma^{t_{tar}^{i''}} \mid s_t=s, a_t=a'' \right ].
    \end{equation}
    Removing $v$ on both sides of Eq.\ref{Q_ineq_final} and according to Eq.\ref{our_Q_traj}, we have
    \begin{equation}
        \tilde{Q}^\pi(s, a', G(s, g)) \geq \tilde{Q}^\pi(s, a'', G(s, g)).
    \end{equation}
    Then,
    \begin{equation}
        \tilde{A}^\pi(s, a', G(s, g)) \geq \tilde{A}^\pi(s, a'', G(s, g)).
    \end{equation}
\end{proof}

Since GCSL aims to mimic the underlying behavior policy using maximum likelihood estimation, DAWOG inherently offers guarantees in relation to GCSL.

\section{Experimental results}

In this section, we examine DAWOG's performance relative to existing state-of-the-art algorithms using environments of increasing complexity. The remainder of this section is organized as follows. The benchmark tasks and datasets are presented in Section \ref{tasks}. The implementation details are provided in Section \ref{implementation}. A list of competing methods is presented in Section \ref{competing_methods}, and the comparative performance results are found in Section \ref{performance}. Here, we also qualitatively inspect the policies learned by DAWOG in an attempt to characterize the improvements that can be achieved over other methods. Section \ref{further_studies} presents extensive ablation studies to appreciate the relative contribution of the different advantage weighting factors. Finally, in Section \ref{sensitivity}, we study how the dual-advantage weight depends on its hyperparameters.  

\subsection{Tasks and datasets} \label{tasks}

\subsubsection{Grid World} 
We designed two $16 \times 16$ grid worlds to assess the performance on a simple navigation task. From its starting position on the grid, an agent needs to reach a goal that has been randomly placed in one of the available cells. Only four actions are available to move left, right, up, and down. The agent accrues a positive reward when it reaches the cell containing the goal. To generate the benchmark dataset, we trained a Deep Q-learning algorithm \cite{mnih2015human}, whose replay buffer, containing $4,000$ trajectories of $50$ time steps, was used as the benchmark dataset.

\subsubsection{AntMaze navigation}

The AntMaze suite used in our experiment is obtained from the D4RL benchmark \cite{fu2020d4rl}, which has been widely adopted by offline GCRL studies \cite{eysenbach2022contrastive,emmons2022rvs,li2022hierarchical}. The task requires to control an 8-DoF quadruped robot that moves in a maze and aims to reach a target location within an allowed maximum of $1,000$ steps. The suite contains three kinds of different maze layouts:  \texttt{umaze} (a U-shape wall in the middle), \texttt{medium} and \texttt{large}, and provides three training datasets. The datasets differ in the way the starting and goal positions of each trajectory were generated: in \texttt{umaze} the starting position is fixed and the goal position is sampled within a fixed-position small region; in \texttt{diverse} the starting and goal positions are randomly sampled in the whole environment; finally, in \texttt{play}, the starting and goal positions are randomly sampled within hand-picked regions. In the sparse-reward environment, the agent obtains a reward only when it reaches the target goal. We use a normalized score as originally proposed in \cite{fu2020d4rl}, i.e.,
\begin{equation*}
s_n = 100 \cdot \frac{s - s_r} { s_e - s_r}
\end{equation*}
where $s$ is the unnormalized score, $s_r$ is a score obtained using a random policy and $s_e$ is the score obtained using an expert policy. 

In our evaluation phase, the policy is tested online. The agent's starting position is always fixed, and the goal position is generated using one of the following methods: 
\begin{itemize}
    \item  \textit{fixed goal}:  the goal position is sampled within a small and fixed region in a corner of the maze, as in previous work \cite{eysenbach2022contrastive,emmons2022rvs,li2022hierarchical}; 
    \item  \textit{diverse goal}: the goal position is uniformly sampled over the entire region. This evaluation scheme has not been adopted in previous works, but helps assess the policy's generalization ability in goal-conditioned settings.  
\end{itemize}

\subsubsection{Gym robotics}

Gym Robotics \cite{plappert2018multi} is a popular robotic suite used in both online and offline GCRL studies \cite{yang2022rethinking, eysenbach2022contrastive}.
The agent to be controlled is a 7-DoF robotic arm, and several tasks are available: in \texttt{FetchReach}, the arm needs to touch a desired location; in \texttt{FetchPush}, the arm needs to move a cube to a desired location;  in \texttt{FetchPickAndPlace} a cube needs to be picked up and moved to a desired location; finally, in \texttt{FetchSlide}, the arm needs to slide a cube to a desired location. Each environment returns a reward of one when the task has been completed within an allowed time horizon of $50$ time steps.  For this suite, we use the expert offline dataset provided by \cite{yang2022rethinking}. The dataset for FetchReach contains  $1 \times 10^5$ time steps whereas all the other datasets contain $2 \times 10^6$ steps. The datasets are collected using a pre-trained policy using DDPG and hindsight relabeling \cite{lillicrap2016continuous, andrychowicz2017hindsight}; the actions from the policy were perturbed by adding Gaussian noise with zero mean and $0.2$ standard deviation.

\subsection{Implementation details} \label{implementation}

DAWOG's training procedure is shown in Algorithm \ref{pseudocode}. In our implementation, for continuous control tasks, we use a Gaussian policy following previous recommendations \cite{raffin2021stablebaselines}. When interacting with the environment, the actions are sampled from the above distribution.  All the neural networks used in DAWOG are 3-layer multi-layer perceptrons with $512$ units in each layer and ReLU activation functions. The parameters are trained using the Adam optimizer \cite{kingma2014adam} with a learning rate $1 \times 10^{-3}$. The training batch size is $512$ across all networks. To represent $G(s, g)$ we use a $K$-dimensional one-hot encoding vector where the $i^{th}$ position is non-zero for the target region and zero everywhere else along with the goal $g$. Four hyper-parameters need to be chosen: the state partition size, $K$, the two coefficients controlling the relative contribution of the two advantage functions, $\beta$ and $\tilde{\beta}$, and the clipping bound, $M$. In our experiments, we use $K=20$ for umaze and medium maze, $K=50$ for large maze, and $K=10$ for all other tasks. In all our experiments, we use fixed values of $\beta = \tilde{\beta} = 10$. The clipping bound is always kept at $M=10$. 

\subsection{Competing methods} \label{competing_methods}

Several competing algorithms have been selected for comparison with DAWOG, including offline DRL methods that were not originally proposed for goal-conditioned tasks and required some minor adaptation. In the remainder of this Section, the nomenclature 'g-' indicates that the original algorithm has been implemented to operate in a goal-conditioned setting by concatenating the state and the goal as a new state and with hindsight relabeling. In all experiments, we independently optimize the hyper-parameters for every algorithm.

The first category of algorithms comprises regression-based methods that imitate the relabeled offline dataset using various weighting strategies: 
\begin{itemize}
    \item GCSL \cite{ghosh2021learning} imitates the relabeled transitions without any weighting strategies.
    \item GEAW \cite{wang2018exponentially, yang2022rethinking} uses goal-conditioned advantage to weight the actions in the offline data.
    \item WGCSL \cite{yang2022rethinking} employs a combination of weighting strategies: discounted relabeling weighting (DRW), goal-conditioned exponential advantage weighting (GEAW), and best-advantage weighting (BAW).
\end{itemize}

We also include three actor-critic methods: 
\begin{itemize}
    \item Contrastive RL \cite{eysenbach2022contrastive} estimates a Q-function by contrastive learning;
    \item g-CQL \cite{kumar2020conservative} learns a conservative Q-function;
    \item g-BCQ \cite{fujimoto2019off} learns a Q-function by clipped double Q-learning with a restricted policy;
    \item g-TD3-BC \cite{fujimoto2021minimalist}  combines TD3 algorithm \cite{fujimoto2018addressing} with a behavior cloning regularizer. 
\end{itemize}
Finally, we include a hierarchical learning method, IRIS \cite{mandlekar2020iris}, which employs a low-level imitation learning policy to reach sub-goals commanded by a high-level goal planner. 

\subsection{Performance comparisons and analysis} \label{performance}

\begin{sidewaystable}
\sidewaystablefn%

\begin{minipage}{\textheight}
\centering
\caption{Experiment results in Gym Robotics.}
\label{exp:gym_res}
\begin{tabular}{l|rrrrrrrrr}
\toprule
Environment        & DAWOG   & GCSL    & WGCSL & GEAW &CRL & g-CQL  & g-TD3. & g-BCQ \\
\midrule
FetchReach        & $\textbf{46.7}_{\pm 0.1}$ & $41.7_{\pm 0.3}$  &$46.3_{\pm 0.1}$& $45.0_{\pm 0.1}$& $46.1_{\pm 0.1}$ &$1.0_{\pm 0.2}$& $45.5_{\pm 0.3}$ & $35.1_{\pm 3.1}$\\
FetchPush         & $\textbf{39.3}_{\pm 0.2}$          & $28.5_{\pm 0.9}$  &$39.1_{\pm 0.2}$ &$37.4_{\pm 0.2}$   & $36.5_{\pm 0.6}$ & $5.7_{\pm 0.8}$ & $30.8_{\pm 0.6}$        & $3.6_{\pm 0.9}$    \\
FetchPickAndPlace &$\textbf{37.9}_{\pm 0.4}$  & $25.2_{\pm 0.8}$  & $34.3_{\pm 0.5}$& $34.5_{\pm 0.5}$& $35.7_{\pm 0.2}$ & $3.2_{\pm 2.5}$& $36.5_{\pm 0.5}$& $1.4_{\pm 0.2}$ \\
FetchSlide        &$9.9_{\pm 0.8}$            & $3.05_{\pm 0.6}$  & $\textbf{10.7}_{\pm 1.0}$ & $4.5_{\pm 1.7}$& $9.9_{\pm 0.2}$& $0.8_{\pm 0.3}$ & $5.8_{\pm 0.6}$& $0.1_{\pm 0.1}$ \\
\bottomrule
\end{tabular}
\end{minipage}

\bigskip 

\begin{minipage}{\textheight}
\centering
\caption{Experiment results in AntMaze environments. The results are normalized by the expert score from D4RL paper. The mean and the standard deviation are calculated by $4$ independent runs.}
\label{exp:antmaze_res}
\begin{tabular}{l|l|rrrrrrrr}
\toprule
            & Environment        & DAWOG   & GCSL    & WGCSL & GEAW & CRL   & g-CQL  & g-TD3. & IRIS\\
\midrule 
            & umaze           & $\textbf{92.8}_{\pm 3.3}$  & $64.4_{\pm 5.0}$  & $85.8_{\pm 2.9}$ & $83.2_{\pm 5.0}$  & $81.9_{\pm 1.7}$  & $66.0_{\pm 2.6}$   & $83.2_{\pm 9.6}$ & $82.6_{\pm 4.7}$\\
            & umaze-diverse    & $\textbf{90.4}_{\pm 3.3}$ & $65.4_{\pm 3.7}$  & $81.6_{\pm 7.8}$ & $70.4_{\pm 5.6}$  & $75.4_{\pm 3.5}$  & $58.4_{\pm 2.4}$   & $77.8_{\pm 8.8}$ & $89.4_{\pm 2.4}$\\
Fixed Goal  & medium-play     & $\textbf{86.6}_{\pm 6.4}$  & $61.8_{\pm 9.2}$  & $50.4_{\pm 7.0}$ & $64.4_{\pm 5.2}$  & $71.5_{\pm 5.2}$  & $27.0_{\pm 4.5}$   & $66.0_{\pm 8.0}$ & $73.1_{\pm 4.5}$\\
            & medium-diverse   & $\textbf{87.4}_{\pm 5.7}$ & $64.2_{\pm 3.7}$  & $46.4_{\pm 6.9}$ & $63.0_{\pm 7.8}$  & $72.5_{\pm 2.8}$  & $32.4_{\pm 4.1}$   & $52.6_{\pm 9.2}$ & $64.8_{\pm 2.6}$\\
            & large-play      & $\textbf{58.6}_{\pm 6.2}$  & $30.6_{\pm 7.6}$  & $30.4_{\pm 6.2}$ & $20.4_{\pm 5.8}$  & $41.6_{\pm 6.0}$  & $8.6_{\pm 4.2}$    & $30.6_{\pm 6.0}$ & $57.9_{\pm 3.6}$\\
            & large-diverse    & $\textbf{52.0}_{\pm 8.6}$ & $32.6_{\pm 8.5}$  & $29.8_{\pm 5.8}$  & $25.2_{\pm 6.7}$  & $49.3_{\pm 6.3}$  & $8.8_{\pm 3.6}$    & $39.8_{\pm 8.9}$ & $43.7_{\pm 1.3}$\\
\midrule
            &umaze           &$\textbf{75.6}_{\pm 5.1}$& $61.2_{\pm 3.1}$&$74.4_{\pm 7.5}$& $67.4_{\pm 3.0}$&$60.4_{\pm 3.2}$&$58.3_{\pm 2.5}$& $63.0_{\pm 5.5}$& -\phantom{---} \\
            &umaze-diverse   &$\textbf{75.2}_{\pm 4.6}$&$58.2_{\pm 5.1}$&$74.0_{\pm 6.3}$&$60.2_{\pm 3.6}$&$49.6_{\pm 4.6}$&$54.6_{\pm 5.2}$&$54.8_{\pm 2.5}$& -\phantom{---} \\
Diverse Goal&medium-play     & $\textbf{70.6}_{\pm 2.6}$ &$45.4_{\pm 5.4}$&$39.2_{\pm 5.2}$&$43.6_{\pm 6.6}$&$26.6_{\pm 2.4}$&$33.6_{\pm 2.6}$& $38.0_{\pm 2.4}$&-\phantom{---} \\
            &medium-diverse  &$\textbf{72.8}_{\pm 3.1}$&$43.2_{\pm 4.3}$&$38.4_{\pm 5.7}$&$49.6_{\pm 5.4}$&$28.6_{\pm 2.8}$&$34.4_{\pm 2.0}$&$45.6_{\pm 5.2}$&-\phantom{---} \\
            &large-play      & $\textbf{44.2}_{\pm 4.3}$ &$35.0_{\pm 6.4}$&$31.4_{\pm 2.4}$&$41.2_{\pm 2.5}$&$24.6_{\pm 5.3}$&$18.3_{\pm 7.5}$&$31.4_{\pm 3.2}$&-\phantom{---} \\
            &large-diverse   & $\textbf{43.8}_{\pm 4.6}$&$35.4_{\pm 2.8}$&$28.3_{\pm 2.8}$&$40.4_{\pm 5.6}$&$26.0_{\pm 2.4}$&$24.6_{\pm 4.1}$&$33.6_{\pm 5.9}$&-\phantom{---} \\
\bottomrule
\end{tabular}
\end{minipage}
\end{sidewaystable}

\begin{figure}[t]
    \centering
    \begin{subfigure}[b]{0.24\textwidth}
    \includegraphics[width=\textwidth]{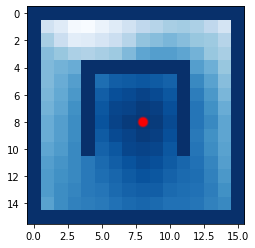} \subcaption{}
    \end{subfigure}
    \begin{subfigure}[b]{0.24\textwidth}
    \includegraphics[width=\textwidth]{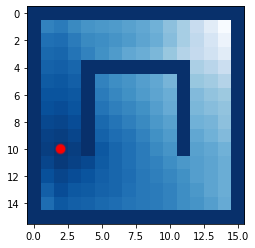} \subcaption{}
    \end{subfigure}
    \begin{subfigure}[b]{0.24\textwidth}
    \includegraphics[width=\textwidth]{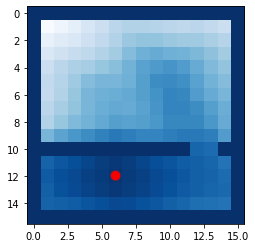} \subcaption{}
    \end{subfigure}
    \begin{subfigure}[b]{0.24\textwidth}
    \includegraphics[width=\textwidth]{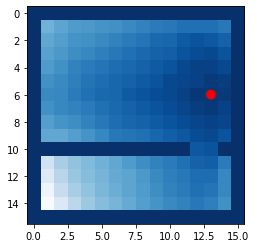} \subcaption{}
    \end{subfigure}
    
    \begin{subfigure}[b]{0.24\textwidth}
    \includegraphics[width=\textwidth]{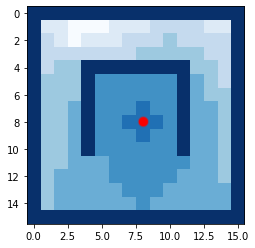} \subcaption{}
    \end{subfigure}
    \begin{subfigure}[b]{0.24\textwidth}
    \includegraphics[width=\textwidth]{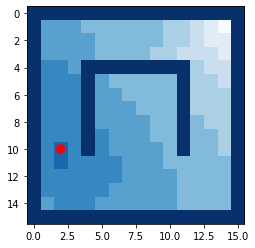} \subcaption{}
    \end{subfigure}
    \begin{subfigure}[b]{0.24\textwidth}
    \includegraphics[width=\textwidth]{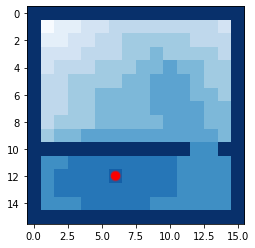} \subcaption{}
    \end{subfigure}
    \begin{subfigure}[b]{0.24\textwidth}
    \includegraphics[width=\textwidth]{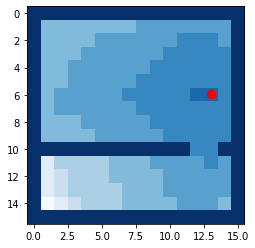} \subcaption{}
    \end{subfigure}
    \caption{An illustration of goal-conditioned state space partitions for two simple Grid World navigation tasks. In each instance, the desired goal is represented by a red circle. In these environments, each state simply corresponds to a position on the grid and, in the top row, is color-coded according to its goal-conditional value. In the lower row, states sharing similar values have been merged to form a partition. For any given state, the proposed target region advantage up-weights actions that move the agent directly towards a neighboring region with higher-value.}
    \label{exp:grid_plot}
\end{figure}

To appreciate how state space partitioning works, we provide examples of valued-based partition for the grid worlds environments in Figure \ref{exp:grid_plot}. In these cases, the environmental states simply correspond to locations in the grid. Here, the state space is divided with darker colors indicating higher values. As expected, these figures clearly show that states can be ordered based on the estimated value function, and that higher-valued states are those close to the goal. We also report the average return across five runs in Table \ref{exp:grid_res} where we compare DAWOG against GCSL and GEAW - two algorithms that are easily adapted for discrete action spaces.  

Table \ref{exp:gym_res} presents the results for the Gym Robotics suite. We detail the average return and the standard deviation for each algorithm, derived from four independent runs with unique seeds. As can be seen from the results, most of the competing algorithms reach a comparable performance with DAWOG. However, DAWOG generally achieves higher scores and the most stable performance in different tasks.

\begin{table}[t]
  \caption{Experiment results for the two Grid World navigation environments.}
  \label{exp:grid_res}
  \small
  \centering
  \begin{tabular}{l|rrr}
    \toprule
    Environment     & DAWOG & GCSL & GEAW \\
    \midrule
    grid-wall       & $\textbf{87.2}_{\pm 1.9}$   
                    & $70.6_{\pm 4.8}$  
                    & $79.8_{\pm 4.0}$ \\
    grid-umaze      & $\textbf{82.4}_{\pm 2.6}$
                    & $68.0_{\pm 3.8}$
                    & $77.2_{\pm 3.5}$\\
    \bottomrule
  \end{tabular}
\end{table}

Table \ref{exp:antmaze_res} displays similar findings for the AntMaze suite. In these more complex, long-horizon environments, DAWOG consistently surpasses all baseline algorithms. In scenarios with diverse goals, while all algorithms exhibit lower performance, DAWOG still manages to secure the highest average score. This setup  requires better generalization performance given that the test goals are sampled from every position within the maze. 

To gain an appreciation for the benefits introduced by the target region approach, in Figure \ref{fig:traj_visual_intro} we visualize $100$ trajectories realized by three different policies for AntMaze tasks: dual-advantage weighting  (DAWOG), equally-weighting and goal-conditioned advantage weighting. The trajectories generated by equally-weighting occasionally lead to regions in the maze that should have been avoided, which results in sub-optimal solutions. The policy from goal-conditioned advantage weighting is occasionally less prone to making the same mistakes, although it still suffers from the multi-modality problem. This can be appreciated, for instance, by observing the \texttt{antmaze-medium} case. In contrast, DAWOG is generally able to reach the goal with fewer detours, hence in a shorter amount of time.   

%The state and the goal are two-dimensional coordinates, so we can visualize the state space partition in Figure \ref{exp:grid_plot}. The experiment results on accumulative returns are shown in Table \ref{exp:grid_res}. 

\subsection{Further studies} \label{further_studies}

In this Section we take a closer look at how the two advantage-based weights featuring in Eq. \ref{eq:weight}  perform, both separately and jointly taken, when used in the loss of Eq. \ref{eq:policy}. We compare learning curves, region occupancy times (i.e. time spent in each region of the state space whilst reaching the goal), and potential overestimation biases. We also study the effects of using different target region and using entropy to regularize policy learning.

\subsubsection{Learning curves} \label{learning_curves}

\begin{figure}[t]
    \centering
    \includegraphics[width=0.96\textwidth]{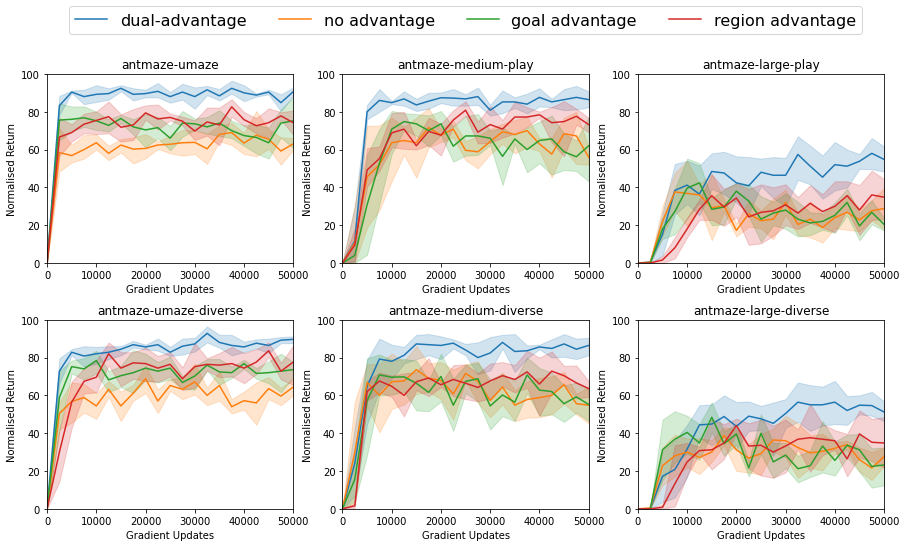}
    \caption{Training curves for different tasks using different algorithms, each one implementing a different weighting scheme: dual-advantage, no advantage, only goal-conditioned advantage, and only the target region advantage. The solid line and the shaded area respectively present the mean and the standard deviation computed from $4$ independent runs.}
    \label{exp:ablation}
\end{figure}

In the AntMaze environments, we train DAWOG using no advantage ($\beta=\tilde{\beta}=0$), only the goal-conditioned advantage ($\beta=10, \tilde{\beta}=0$), only the target region advantage ($\beta=0, \tilde{\beta}=10$), and the proposed dual-advantage ($\beta=\tilde{\beta}=10$). 
Over the course of $50,000$ gradient updates, Figure \ref{exp:ablation} clearly illustrates the distinct learning trajectories of each algorithm. Both the goal-advantage and region-based advantage perform better than using no advantage, and their performance is generally comparable, with the latter often achieving higher normalized returns. Combining the two advantages leads to significantly higher returns than any advantage weight individually taken. 

\subsubsection{Region occupancy times}

\begin{figure}[t]
    \centering
    \includegraphics[width=0.45\textwidth]{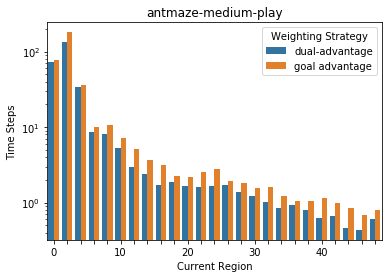}
    \includegraphics[width=0.45\textwidth]{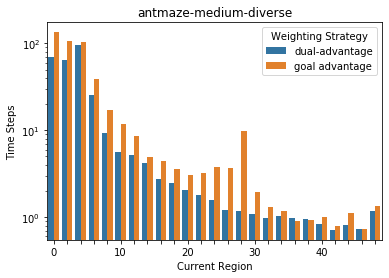}
    \includegraphics[width=0.45\textwidth]{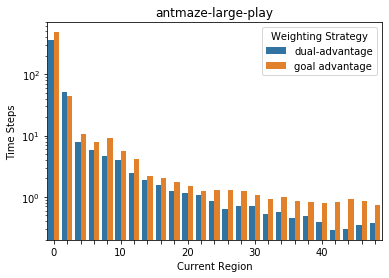}
    \includegraphics[width=0.45\textwidth]{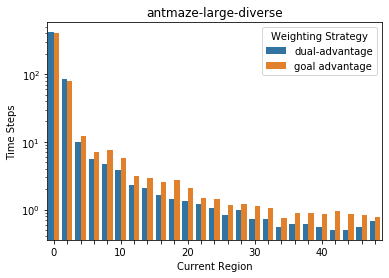}
    \caption{ Average time spent in a region of the state space before moving on to the higher-ranking region ($K=50$) using a goal-conditioned value function for state partitioning. The $y$-axis indicates the average number of time steps (in log scale) spent in a region. The dual-advantage weighting scheme allows the agent to reach each subsequent target region more rapidly compared to the goal-conditioned advantage alone, which results in overall shorter time spent to reach the final goal.}
    \label{fig:steps_to_target_region}
\end{figure}

In this study, we set out to confirm that the dual-advantage weighting scheme results in a policy favoring actions leading to the next higher ranking target region rapidly, i.e. by reducing the occupancy time in each region.  Using the AntMaze environments, Figure \ref{fig:steps_to_target_region} shows the average time spent in a region of the state space partitioned with $K=50$ regions.  As shown here, the dual-advantage weighting allows the agent to reach the target (next) region in fewer time steps compared to the goal-conditioned advantage alone. As the episode progresses, the ant's remaining time to complete its task diminishes, influencing its decision-making process. Hence, as the ant progressively moves to higher ranking regions closer to the goal, the occupancy times decrease.  

\subsubsection{Over-estimation bias}  \label{bias}

\begin{figure}[t]
    \centering
    \includegraphics[width=0.9\textwidth]{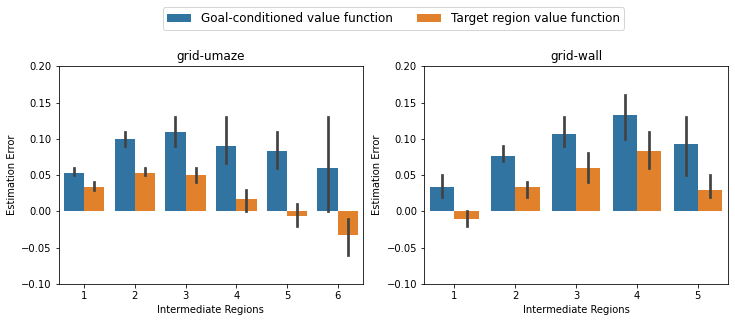}
    \caption{ Estimation error of goal-conditioned and target region value functions in Grid World tasks.}
    \label{exp:estimation_bias}
\end{figure}

We assess the extent of potential overestimation errors affecting the two advantage weighting factors used in our method (see Eq. \ref{eq:weight}). This is done by studying the error that occurred in the estimation of the corresponding V-functions (see Eq. \ref{goal-advantage} and Eq. \ref{region-advantage}). Given a state $s$ and goal $g$, we compute the goal-conditioned V-value estimation error as $V_{\psi_1}(s, g) - V^{\pi}(s, g)$, where $V_{\psi_1}(s, g)$ is the parameterized function learned by our algorithm and  $V^{\pi}(s, g)$ is an unbiased Monte-Carlo estimate of the goal-conditioned V-function's true value \cite{sutton2018reinforcement}. Since $V^{\pi}(s, g)$ represents the expected discounted return obtained by the underlying behavior policy that generates the relabeled data, we use a policy pre-trained with the GCSL algorithm to generate $1,000$ trajectories to calculate the Monte-Carlo estimate (i.e. the average discounted return). Analogously, the target region V-value estimation error is $\tilde{V}_{\psi_2}(s, g, G(s, g)) - \tilde{V}^{\pi}(s, g, G(s,g))$. We use the learned target region V-value function to calculate $\tilde{V}_{\psi_2}(s, g, G(s, g))$, and Monte-Carlo estimation to approximate $\tilde{V}^{\pi}(s, g, G(s,g))$. 

Our investigation focuses on the Grid World environment, specifically analyzing two distinct layouts: grid-umaze and grid-wall.
For each layout, we randomly sample $s$ and $g$ uniformly within the entire maze and ensure that the number of regions separating them is uniformly distributed in $\{1, \ldots, K\}$. Then, for each $k$ in that range: 1) $1,000$ goal positions are sampled randomly within the whole layout; 2) for each goal position, the state space is partitioned according to $V_{\psi}(\cdot,g)$; and 3) a state is sampled randomly within the corresponding region. Since there may exist regions without any states, the observed total number of regions is smaller than $K=10$. The resulting estimation errors are shown in Fig. \ref{exp:estimation_bias}. As can be seen here, both the mean and standard deviation of the $\tilde{V}$-value errors are consistently smaller than those corresponding to the $V$-value errors. This indicates the target region value function is more robust against over-estimation bias, which may help improve the generalization performance in out-of-distribution settings.

\subsubsection{Effects of different target regions}

\begin{figure}[t]
    \centering
    \includegraphics[width=0.4\textwidth]{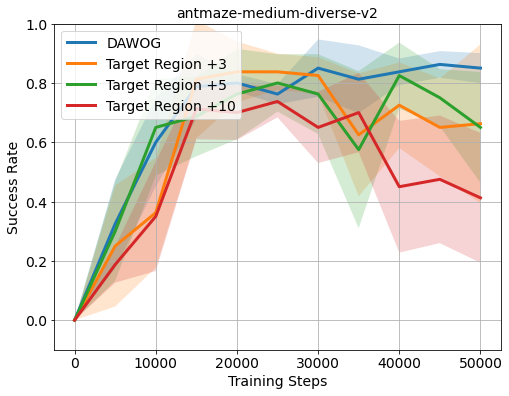}
    \includegraphics[width=0.4\textwidth]{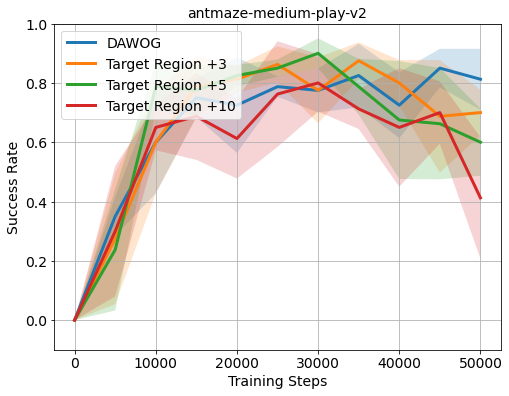}
    \includegraphics[width=0.4\textwidth]{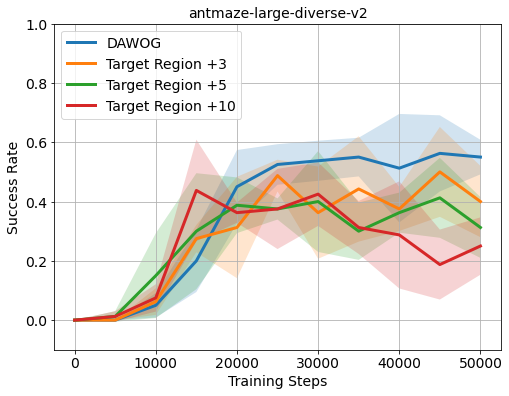}
    \includegraphics[width=0.4\textwidth]{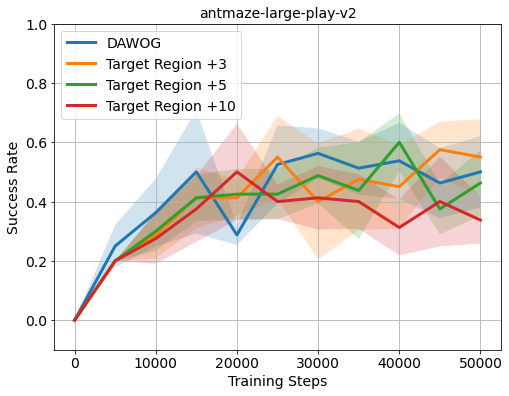}
    \caption{DAWOG with different target regions. The target region is set to be the next (DAWOG), $3$, $5$, and $10$ after the current region. Each curve in the graph is generated from four distinct runs, each initiated with a different random seed.}
    \label{fig:different_targets}
\end{figure}

As outlined in Definition \ref{def:target_region}, the target region comprises states with goal-conditioned values marginally exceeding the current state's value. Within DAWOG, when the current region is denoted as $B_k(g)$, the subsequent target region becomes $B_{k+1}(g)$. Yet, for immediate benefits, regions beyond $B_{k+1}(g)$ can also be contemplated. This section delves into the implications of varying target regions. Figure \ref{fig:different_targets} demonstrates that targeting the immediate neighboring region with higher values using DAWOG consistently yields superior performance compared to other configurations with varied target regions. There is only one instance (antmaze-large-play) where targeting a slightly further region yields a marginally better outcome. Nonetheless, as a general trend, the performance advantage diminishes as the target region becomes increasingly distant from the current region.

\subsubsection{Policy learning with entropy regularization} \label{sec:regularization}

\begin{figure}[t]
    \centering
    \includegraphics[width=0.4\textwidth]{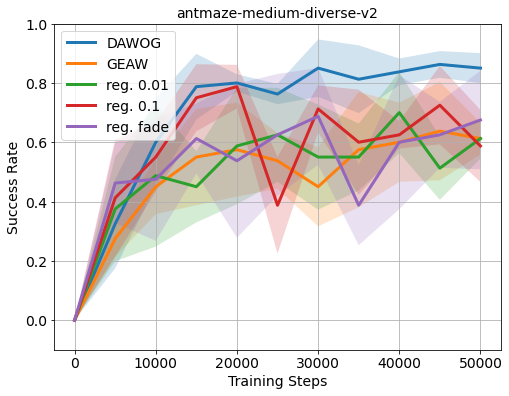}
    \includegraphics[width=0.4\textwidth]{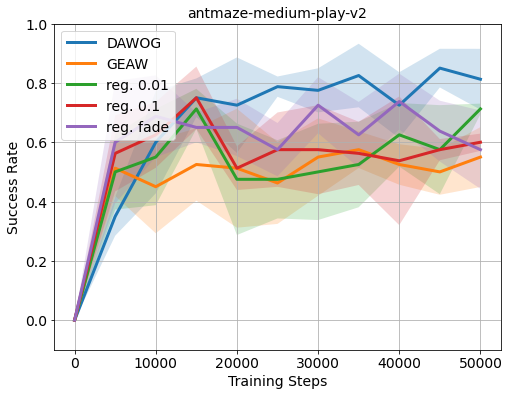}
    \includegraphics[width=0.4\textwidth]{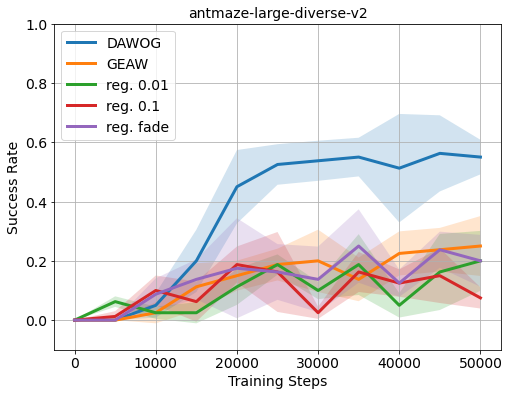}
    \includegraphics[width=0.4\textwidth]{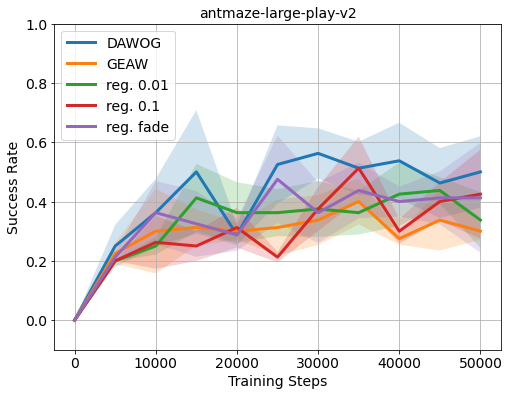}
    \caption{GEAW with entropy regularization. The plots show DAWOG, GEAW with regularization $\alpha=0$, $\alpha=0.1$, $\alpha=0.01$, and $\alpha$ gradually decreasing from $0.1$ to $0.01$ (fade). All curves are obtained by $4$ different random seeds.}
    \label{fig:entropy_regularization}
\end{figure}

The concept of target region advantage can be perceived as a regularization technique. In this analysis, we juxtapose DAWOG with a version of GEAW enhanced by entropy regularization. The refined objective function is expressed as:
\begin{equation*}
J_{reg}(\pi) = \mathbb{E}_{(s_t, a_t, g) \sim \mathcal{D}_R} \left [ \exp_{clip} (\beta A^{\pi_b}(s_t, a_t, g)) \log \pi(a_t \mid s_t, g) + \alpha \mathcal{H}(\pi(\cdot \mid s_t, g))\right ]
\end{equation*}
where $\mathcal{H}(\pi(\cdot \mid s_t, g))$ is defined as $\frac{1}{2} \ln 2\pi e \sigma^2$, with $\sigma$ representing the standard deviation of the Gaussian distribution conditioned as $\pi(\cdot \mid s_t, g)$. Initially, we set $\alpha$ values from the set $0$, $0.01$, $0.1$. Subsequently, we employ a dynamic approach for $\alpha$, allowing it to decrease progressively from $0.1$ to $0.01$. The outcomes of these experiments are depicted in Figure \ref{fig:entropy_regularization}. 

Although strategically adjusted regularization can slightly improve the GEAW baseline, it is evident that DAWOG maintains a consistent edge in performance. This superior performance of DAWOG can be attributed to the unique manner in which it introduces short-term goals.

\subsection{Sensitivity to hyperparameters} \label{sensitivity}

\begin{figure}[t]
    \centering
    \includegraphics[width=0.60\linewidth]{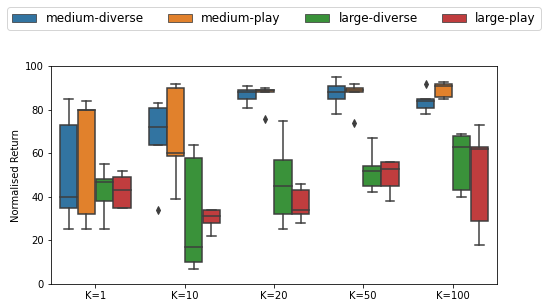}
    \bigskip
    \caption{DAWOG's performance, evaluated on the AntMaze dataset, as a function of $K$, the number of state space partitions required to define the target regions. The box plot of the normalized return in AntMaze task is achieved by DAWOG in four settings when the target region size decreases ($K$ increases). We used $4$ runs with different seeds. Best performance (highest average returns and lowest variability) was consistently achieved across all settings with around $K=50$ equally sized target regions.}
    \label{exp:diff_k}
\end{figure}

\begin{figure}[t]
    \centering
    \includegraphics[width=0.9\linewidth]{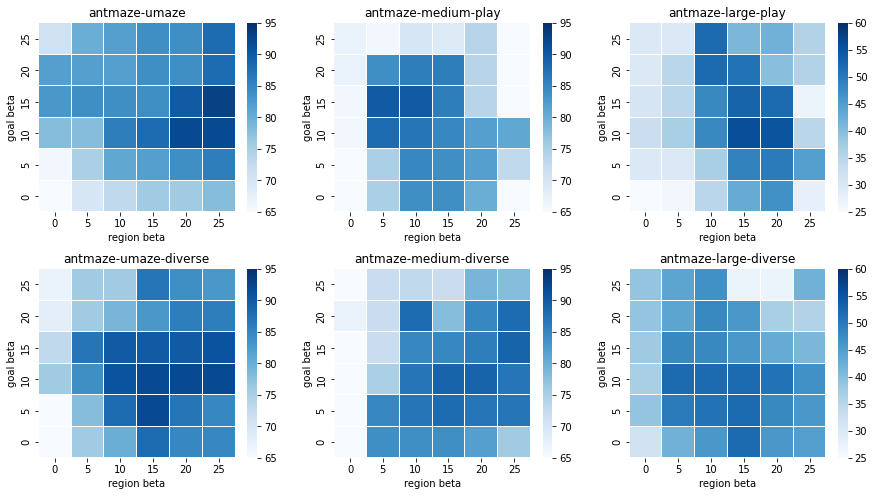}
    \bigskip
    \caption{DAWOG's performance, evaluated on six datasets, as a function of its two hyperparameters, $\beta$ and $\tilde{\beta}$, controlling the goal-conditioned and target-based exponential weights featuring in Equation \ref{eq:weight}. The performance metric is the average return across 5 runs. To produce the results presented in Table \ref{exp:antmaze_res}, we used a (potentially sub-optimal) fixed parameter combination: $\beta=10$ and $\tilde{\beta}=10$.}
    \label{exp:beta_study}
\end{figure}

Lastly, we examine the impact of the number of partitions ($K$) and the coefficients $\beta$ and $\tilde{\beta}$, which control the relative contribution of the two advantage functions on DAWOG's overall performance. In the AntMaze task, we report the distribution of normalized returns as $K$ increases. Figure \ref{exp:diff_k} reveals that an optimal parameter yielding high average returns with low variance often depends on the specific task and is likely influenced by the environment's complexity.

The performance of DAWOG, as depicted in Figure \ref{exp:beta_study}, varies in response to different settings of $\beta$ and $\tilde{\beta}$, highlighting the algorithm's sensitivity to these parameters. The plot demonstrates some minimal sensitivity to various parameter combinations but also exhibits a good degree of symmetry. In all our experiments, including those in Tables \ref{exp:antmaze_res} and \ref{exp:gym_res}, we opted for a shared value, $\beta=\tilde{\beta}=10$, rather than optimizing each parameter combination for each task. This choice suggests that strong performance can be achieved even without extensive hyperparameter optimization.

\section{Discussion and conclusions}

Our study introduces a novel dual-advantage weighting scheme in supervised learning, specifically designed to tackle the complexities of multi-modality and distribution shifts in goal-conditioned offline reinforcement learning (GCRL). The corresponding algorithm, DAWOG (Dual-Advantage Weighting for Offline Goal-conditioned learning), prioritizes actions that lead to higher-reward regions, introducing an additional source of inductive bias and enhancing the ability to generalize learned skills to novel goals. Theoretical support is provided by demonstrating that the derived policy is never inferior to the underlying behavior policy. Empirical evidence shows that DAWOG learns highly competitive policies and surpasses several existing offline algorithms on demanding goal-conditioned tasks. Significantly, the ease of implementing and training DAWOG underscores its practical value, contributing substantially to the evolving understanding of offline GCRL and its interplay with goal-conditioned supervised learning (GCSL).

The potential for future research in refining and expanding upon our proposed approach is multifaceted. Firstly, our current method partitions states into equally-sized bins for the value function. Implementing an adaptive partitioning technique that does not assume equal bin sizes could provide finer control over state partition shapes (e.g., merging smaller regions into larger ones), potentially leading to further performance improvements.

Secondly, considering DAWOG's effectiveness in alleviating the multi-modality problem in offline GCRL, it may also benefit other GCRL approaches beyond advantage-weighted GCSL. Specifically, our method could extend to actor-critic-based offline GCRL, such as TD3-BC \cite{fujimoto2021minimalist} , which introduces a behavior cloning-based regularizer into the TD3 algorithm \cite{fujimoto2018addressing} to keep the policy closer to actions experienced in historical data. The dual-advantage weighting scheme could offer an alternative direction for developing a TD3-based algorithm for offline GCRL.

Lastly, given our method's ability to accurately weight actions, it might also facilitate exploration in online GCRL, potentially in combination with self-imitation learning  \cite{oh2018self, ferret2021self, li2022phasic}. For example, a recent study demonstrated that advantage-weighted supervised learning is a competitive method for learning from good experiences in GCRL settings \cite{li2022phasic}. These promising directions warrant further exploration.

\section*{Acknowledgments}

Giovanni Montana acknowledges support from a UKRI AI Turing Acceleration Fellowship (EPSRC EP/V024868/1).

\pagebreak

%\bibliography{references}% common bib file


\appendix

\pagebreak

\section*{Appendix}

\renewcommand{\thesubsection}{\Alph{subsection}.}

In this section, we show an example to motivate our algorithm and provide more evidence to support the argument in the main paper.

\subsection{A numerical example} \label{sec:motivating_example}
\begin{figure}[ht]
            \centering

            \begin{subfigure}{0.48\textwidth}
                \includegraphics[width=0.8\textwidth]{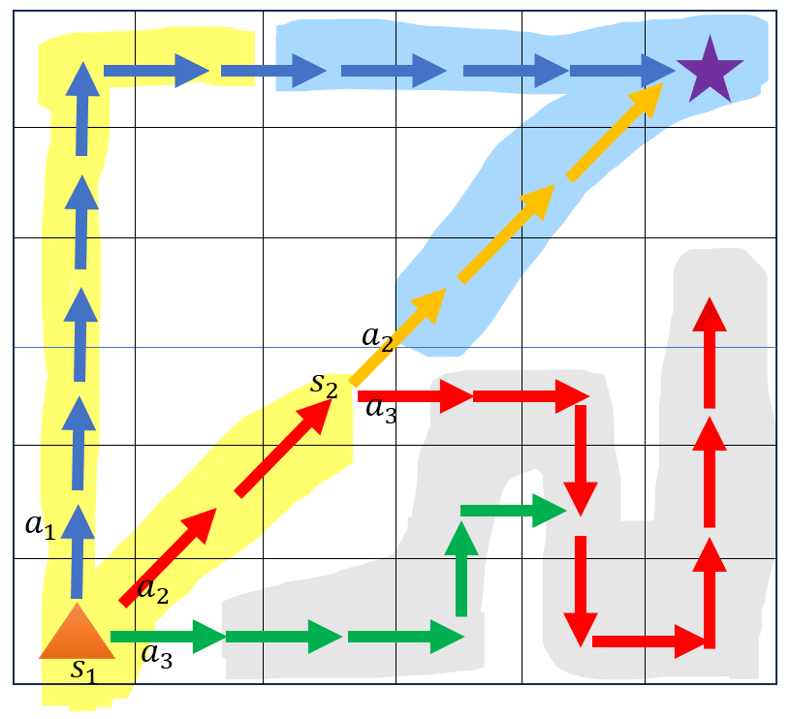}        
                \subcaption{}
            \end{subfigure}
            \begin{subfigure}{0.48\textwidth}
                \includegraphics[width=0.8\textwidth]{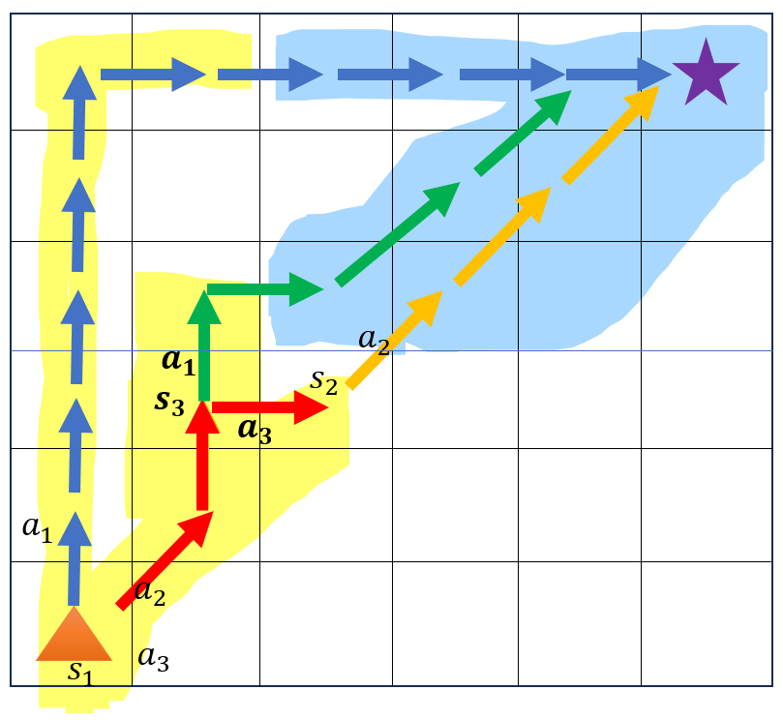}        
                \subcaption{}
            \end{subfigure}
            \caption{A motivating example on a grid environment.}
            \label{fig:toy_example}
        \end{figure}
        
In this section, we use numerical examples to highlight the advantages of our dual-advantage weighting scheme over the GEAW approach. At its core, our method introduces the concept of 'region-advantage'. This involves segmenting states into ordered regions based on ascending state values. By doing so, we can harness both short-term and traditional goal-conditioned advantages to more effectively weight different behavioral actions. A key strength of our approach is its ability to identify the most effective action towards a goal using the region-advantage value. This is something the GEAW approach might miss, especially when the optimal action doesn't possess the highest goal-conditioned advantage value due to the complexities of action multi-modality.

In Figure~\ref{fig:toy_example} (a), four behavioral trajectories are depicted, originating from the starting state (represented by an orange triangle) and aiming for the goal (indicated by a purple star). Due to constraints on time, only two of these trajectories successfully reach the goal. At the starting state, $s_1$, three potential actions are available: $a_1$, $a_2$, and $a_3$. The goal-conditioned advantage values for these actions can be determined through the subsequent calculations.

Given the reward function that only gives $r=1$ at the goal state otherwise $r=0$, and discount factor $\gamma=0.99$. The goal-conditioned state-action values for $s_2, g$ and $a_2, a_3$ are:
        \begin{equation}
        \begin{array}{ll}
            & Q^{\pi_b}(s_2, a_2, g) = 0.99^3  \\
            & Q^{\pi_b}(s_2, a_3, g) = 0   \\
        \end{array}
        \end{equation}
        The goal-conditioned state value of $s_2$:
        \begin{equation}
            V^{\pi_b}(s_2, g) = \frac{0.99^3 + 0}{2} \approx 0.485
        \end{equation}
        The goal-conditioned state-action value for $s_1, g$ and $a_1, a_2$:
        \begin{equation}
        \begin{array}{ll}
            & Q^{\pi_b}(s_1, a_1, g) = 0.99^{10} \approx 0.904  \\
            & Q^{\pi_b}(s_1, a_2, g) = 0.99^2 \cdot 0.485 \approx 0.475  \\
            % & Q^{\pi_b}(s_1, a_3, g) = 0   \\
        \end{array}
        \end{equation}
        % The goal-conditioned state value of the state $s_1$ is 
        % \begin{equation}
        %     V^{\pi_b}(s_1, g) \approx \frac{0.904 + 0.475 + 0}{3} \approx 0.460
        % \end{equation}
        According to $A^{\pi_b}(s, a, g) = Q^{\pi_b}(s, a, g) - V^{\pi_b}(s)$, we have:
        \begin{equation}
        \begin{array}{ll}
            % & A^{\pi_b}(s_1, a_1, g) \approx 0.449  \\
            % & A^{\pi_b}(s_1, a_2, g) \approx 0.005  \\
            % & A^{\pi_b}(s_1, a_3, g) \approx -0.455   \\
            A^{\pi_b}(s_1, a_1, g) > A^{\pi_b}(s_1, a_2, g)
        \end{array}
        \end{equation}
        
According to the state values, the states can be roughly divided into three regions colored by blue, yellow, and gray. For $s_1$, the next region is the blue one (the states in this region have bigger state values). 
The target region $Q-$values of $s_1, G$, and $a_1, a_2$ are:
        \begin{equation}
        \begin{array}{ll}
            & \tilde{Q}^{\pi_b}(s_1, a_1, G) = 0.99^7 \approx 0.932  \\
            & \tilde{Q}^{\pi_b}(s_1, a_2, G) = 0.99^3 \approx 0.970  
        \end{array}
        \end{equation}
        Thus, we have:
        \begin{equation}
            \tilde{A}^{\pi_b}(s_1, a_1, G) < \tilde{A}^{\pi_b}(s_1, a_2, G).
        \end{equation}
    
From the example, it's evident that $a_2$ is the optimal action towards the goal. However, due to the trajectory's failure post $s_2$, its goal-conditioned advantage value, $A^{\pi_b}(s_1,a_2, g)$, is less than that of $A^{\pi_b}(s_1,a_1, g)$. In contrast, our region-advantage value focuses on the action's advantage towards the subsequent region. This provides a nuanced, short-term assessment of an action's quality, enabling a more accurate identification of optimal actions compared to relying solely on the goal-conditioned advantage value.

Figure~\ref{fig:toy_example} (b) underscores the advantages of integrating both goal-conditioned and region-conditioned advantages. In this illustration, state $s_3$ presents two actions, both of which exhibit identical region-advantage values: $\tilde{A}^{\pi_b}(s_1, a_1, G) = \tilde{A}^{\pi_b}(s_1, a_3, G)$. Additionally, the goal-conditioned advantage values reveal that $A^{\pi_b}(s_3, a_1, g) > A^{\pi_b}(s_3, a_3, g)$. This suggests that blending goal-conditioned and region-conditioned advantages for re-weighting behavioral actions potentially offers enhanced outcomes compared to solely employing the region-conditioned advantage.

It is important to note that the numerical examples in this section serve primarily as illustrative tools. A more comprehensive and rigorous exploration, encompassing both theoretical and empirical analyses, is detailed in the main body of the paper.

\pagebreak

\subsection{Assessing GEAW with two value functions} \label{sec:ensemble}

\begin{figure}[t]
    \centering
    \includegraphics[width=0.4\textwidth]{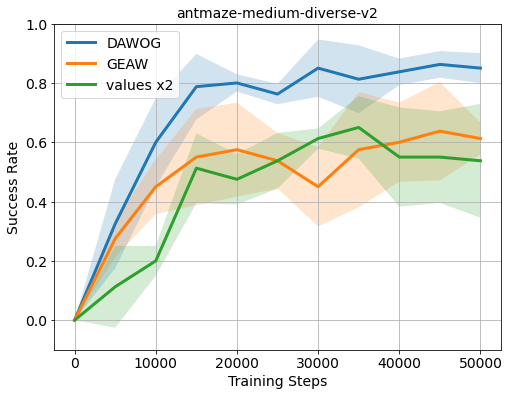}
    \includegraphics[width=0.4\textwidth]{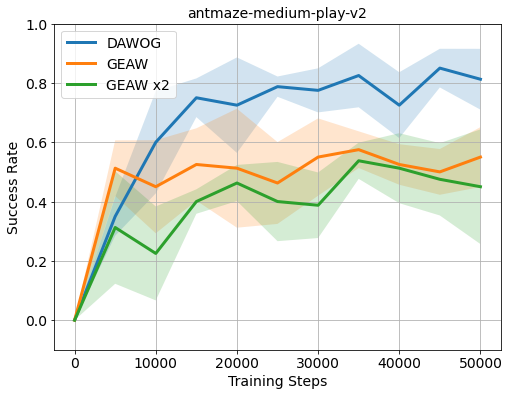}
    \includegraphics[width=0.4\textwidth]{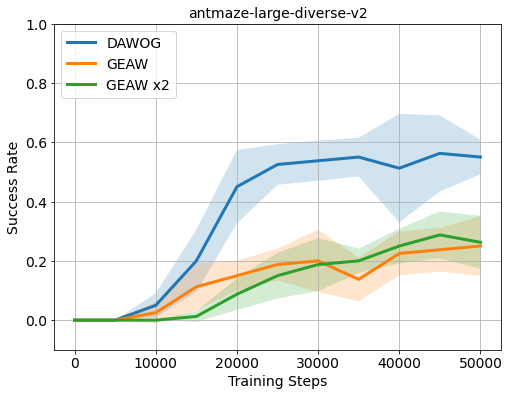}
    \includegraphics[width=0.4\textwidth]{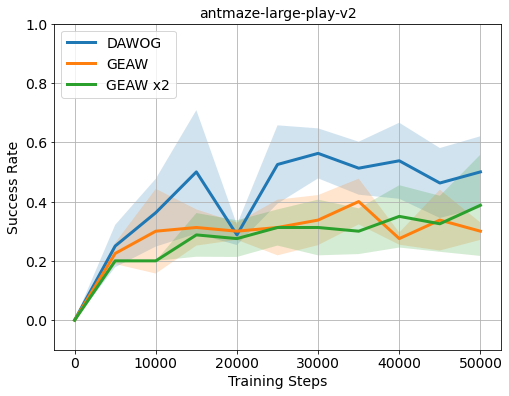}
  \caption{The plots depict the performance of DAWOG, standard GEAW, and GEAW utilizing two value functions (denoted as GEAW x2) across four environments. Each curve represents an average derived from $4$ distinct random seeds.}
    \label{fig:ensemble_geaw}
\end{figure}

In this experiment, we aim to ascertain if the performance of DAWOG is primarily due to the employment of two value networks with unique initializations. To test this, we substituted DAWOG's target region advantage with a goal-conditioned advantage. Specifically, we initialized two goal-conditioned state value functions, represented as \( \{V_{\theta_i}\}^2_{i=1} \), using distinct random seeds. These were then updated following the GEAW protocol. The policy was optimized using
$$
    J(\pi) = \mathbb{E}_{(s_t, a_t, r_t, g) \sim \mathcal{D}_R} \left [ \exp_{clip} (\frac{\beta}{2} \sum^2_{i=1} A^{\pi_b}_i(s_t, a_t, g)) \log \pi(a_t \mid s_t, g) \right ]
    $$
where the advantage is calculated as $A^{\pi_b}_i(s_t, a_t, g)= r_t + \gamma V_{\theta_i}(s_{t+1}, g) - V_{\theta_i}(s_t, g)$. We benchmarked this approach against both GEAW and DAWOG across four environments, as depicted in Figure \ref{fig:ensemble_geaw}. Our findings did not indicate that DAWOG's superior performance is solely due to the dual value networks with different initializations.

\pagebreak

\subsection{Assessing the effects of DAWOG's target region} \label{sec:success}

\begin{figure}[t]
    \centering
    \includegraphics[width=0.4\textwidth]{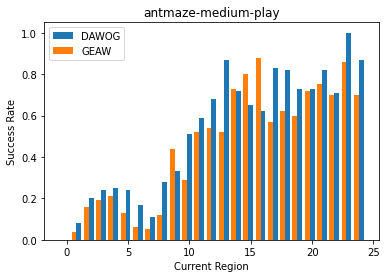}
    \includegraphics[width=0.4\textwidth]{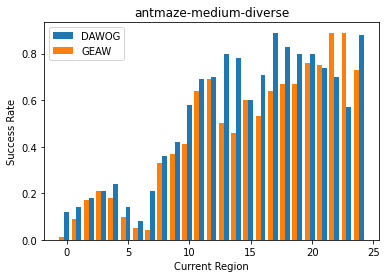}
    \includegraphics[width=0.4\textwidth]{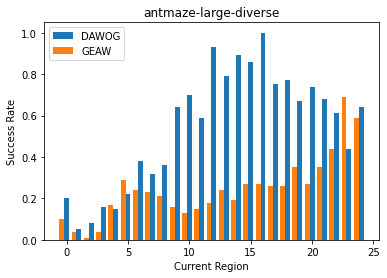}
    \includegraphics[width=0.4\textwidth]{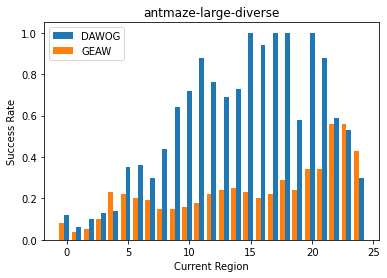}
   \caption{Success rate comparison of two weighting strategies in reaching the subsequent target region. Here, the success rate indicates the likelihood of the agent successfully reaching the subsequent target region within a span of ten time steps.}
    \label{fig:success}
\end{figure}

In this subsection, we provide further evidence highlighting the efficacy of the dual-advantage in promoting short-term success for the agent. Figure \ref{fig:success} depicts the success rate at which the agent reaches the subsequent target region within a span of ten time steps. When comparing GEAW with DAWOG, it becomes evident that by harnessing the target region advantage, the agent more frequently reaches the target region in fewer steps, thereby progressing closer to the intended goal.

\pagebreak

% \backmatter

\pagebreak

\end{document}